\documentclass{article}

\usepackage{amsmath,amsfonts,bm}

\def\eqref#1{equation~\ref{#1}}

\def\1{\bm{1}}

\DeclareMathAlphabet{\mathsfit}{\encodingdefault}{\sfdefault}{m}{sl}
\SetMathAlphabet{\mathsfit}{bold}{\encodingdefault}{\sfdefault}{bx}{n}

\usepackage{hyperref}
\usepackage{url}            
\usepackage{booktabs}       
\usepackage{amsfonts}       
\usepackage{nicefrac}       
\usepackage{microtype}      
\usepackage{algorithm}
\usepackage{algorithmic}
\usepackage{amsmath}
\usepackage{times}
\usepackage{mathtools}
\usepackage[numbers]{natbib}
\usepackage{color}
\usepackage{siunitx}
\usepackage{multirow}
\usepackage{amsthm}
\usepackage{mathrsfs}
\usepackage{graphicx}
\usepackage{caption}
\usepackage{xspace}
\usepackage[export]{adjustbox}
\usepackage{float}
\usepackage{enumitem}
\usepackage{listings}
\usepackage{wrapfig}
\usepackage{amsthm}
\usepackage{subcaption}
\usepackage{wrapfig}
\usepackage{xcolor}
\usepackage{footnote}
\usepackage{cleveref}
\usepackage{lipsum}

\usepackage[compact]{titlesec}

\usepackage{hyperref}
\usepackage{cleveref}
\usepackage{url}

\usepackage[accepted]{icml2021}

\icmltitlerunning{Function Contrastive Learning of Transferable Meta-Representations}

\begin{document}

\twocolumn[
\icmltitle{Function Contrastive Learning of Transferable Meta-Representations}

\icmlsetsymbol{equal}{*}

\begin{icmlauthorlist}
\icmlauthor{Muhammad Waleed Gondal}{mpi}
\icmlauthor{Shruti Joshi}{mpi}
\icmlauthor{Nasim Rahaman}{mpi,mila}
\icmlauthor{Stefan Bauer}{mpi,ed}
\icmlauthor{Manuel W\"uthrich}{mpi}
\icmlauthor{Bernhard Sch\"olkopf}{mpi}
\end{icmlauthorlist}

\icmlaffiliation{mpi}{Max Planck Institute for Intelligent Systems, T\"ubingen, Germany}
\icmlaffiliation{mila}{Mila, University of Montreal, Montreal, Canada}
\icmlaffiliation{ed}{CIFAR Azrieli Global Scholar}

\icmlcorrespondingauthor{Muhammad Waleed Gondal}{waleed.gondal@tue.mpg.de}

\icmlkeywords{Machine Learning, ICML}

\vskip 0.3in
]

\printAffiliationsAndNotice{} %

\begin{abstract}

Meta-learning algorithms adapt quickly to new tasks that are drawn from the same task distribution as the training tasks. The mechanism leading to fast adaptation is the conditioning of a downstream predictive model on the inferred representation of the task's underlying data generative process, or \emph{function}. This \emph{meta-representation}, which is computed from a few observed examples
of the underlying function, is learned jointly with the predictive model. In this work, we study the implications of this joint training on the transferability of the meta-representations. Our goal is to learn meta-representations that are robust to noise in the data and facilitate solving a wide range of downstream tasks that share the same underlying functions. To this end, we propose a decoupled encoder-decoder approach to supervised meta-learning, where the encoder is trained with a contrastive objective to find a good representation of the underlying function. In particular, our training scheme is driven by the self-supervision signal indicating whether two sets of examples
stem from the same function. Our experiments on a number of synthetic and real-world datasets show that the representations we obtain outperform strong baselines in terms of downstream performance and noise robustness, even when these baselines are trained in an end-to-end manner.
\end{abstract}

\section{Introduction}
\label{intro}

Many supervised learning problems are concerned with approximating a data-generating function $f : \mathcal{X} \to \mathcal{Y}$ given a finite set of $N$ samples, $\{x_i, y_i = f(x_i)\}_{i=1}^{N}$. 
Expressive models, such as deep neural networks, are known to excel at this function approximation task, but they often heavily rely on the number of samples $N$ being large. This poses further challenges: in many domains of interest, sourcing enough data is a challenging endeavour; further, the process of training such models can be prohibitively slow for many applications. 
This is exacerbated by the fact that in the typical setting, each new data-generating function encountered requires that the model be retrained. In other words, the model is not shared between data-generating functions, even when training a model to approximate one function can potentially be beneficial for approximating another function.

To overcome these challenges, a variety of meta-learning methods have been proposed \citep{vinyals2016matching, snell2017prototypical, garnelo2018conditional, ravi2016optimization, finn2017model}. 
In the present work, we are interested in a class of models that use encoder-decoder architectures such as Conditional Neural Processes (CNPs) \citep{garnelo2018conditional} and Generative Query Networks (GQNs) \citep{eslami2018neural}. In the first stage, an encoder is used to infer a fixed-dimensional representation of a given function $f$ from just a few input-output examples $O^f=\{(x_i, y_i)\}_i$, \emph{the context dataset}.
We call it the \emph{meta-representation} of the function $r = h_{\phi} (O^f)$, where $h$ is an encoder parameterized by $\phi$. In the second stage, the meta-representation is then used to condition a predictive model in order to solve a downstream prediction task related to that function. For instance, the task may consist of predicting the function value $y$ at unseen locations $x$ or classifying images after observing only a few pixels (in that case, $x$ is the pixel location and $y$ is the pixel value). This two-stage process has multiple benefits. First, the extraction of prior knowledge about $f$ directly from the training data, in the form of meta-representation, reduces the need for specifying inductive biases (model architectures, training details, etc.) particular to $f$. Thus, it allows learning to be shared between functions such that a single model can be trained on a distribution over functions. Second, the computation of meta-representations is efficient and can be done online. Third, the computation of meta-representations provides flexibility to solve a variety of downstream tasks concerning a specific function.

\begin{figure*}%
\begin{subfigure}{0.33\textwidth}
\centering\includegraphics[width=\textwidth]{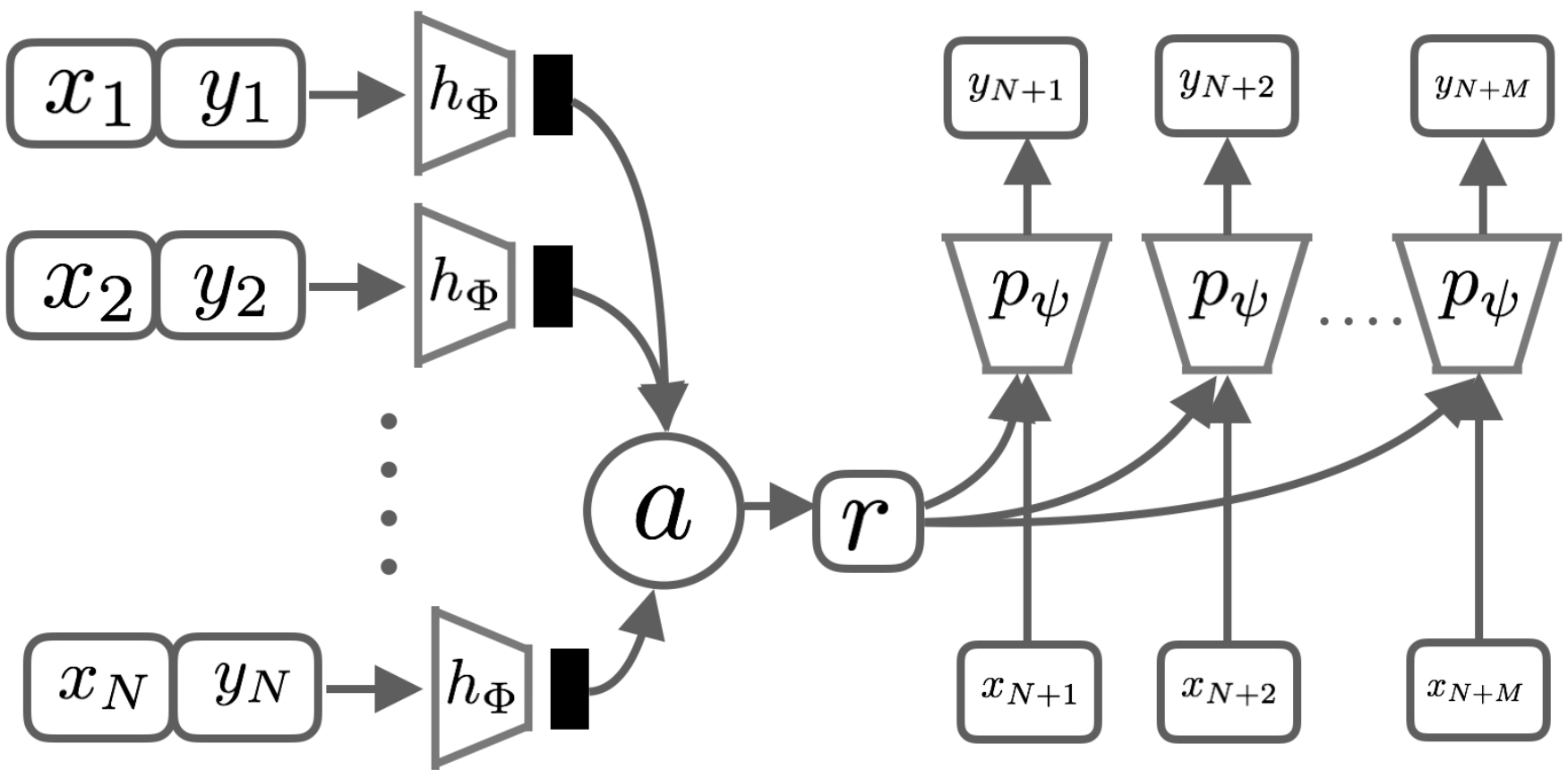}
\caption{CNP}
\end{subfigure}\hspace{5mm} %
\begin{subfigure}{0.24\textwidth}
\centering\includegraphics[width=\textwidth]{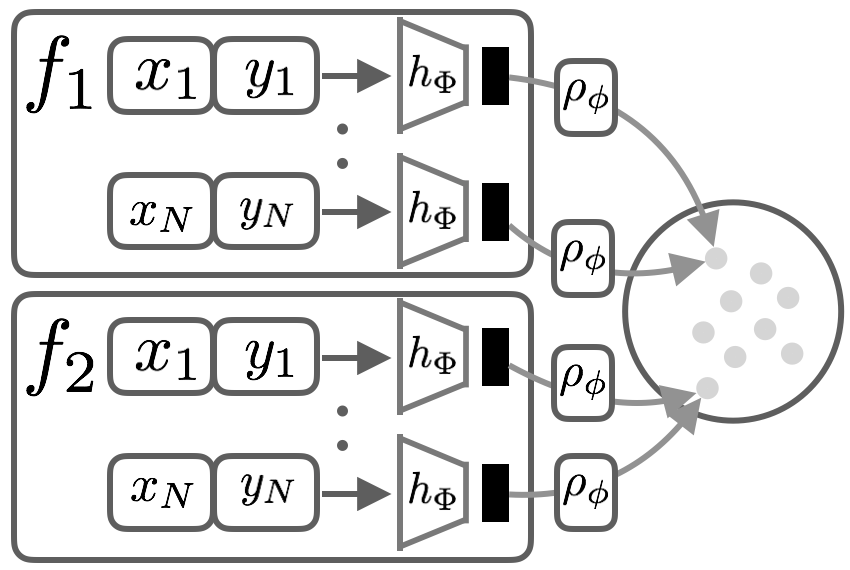}
\caption{FCRL (encoder training)}
\end{subfigure}\hspace{5mm}
\begin{subfigure}{0.38\textwidth}
\centering\includegraphics[width=\textwidth]{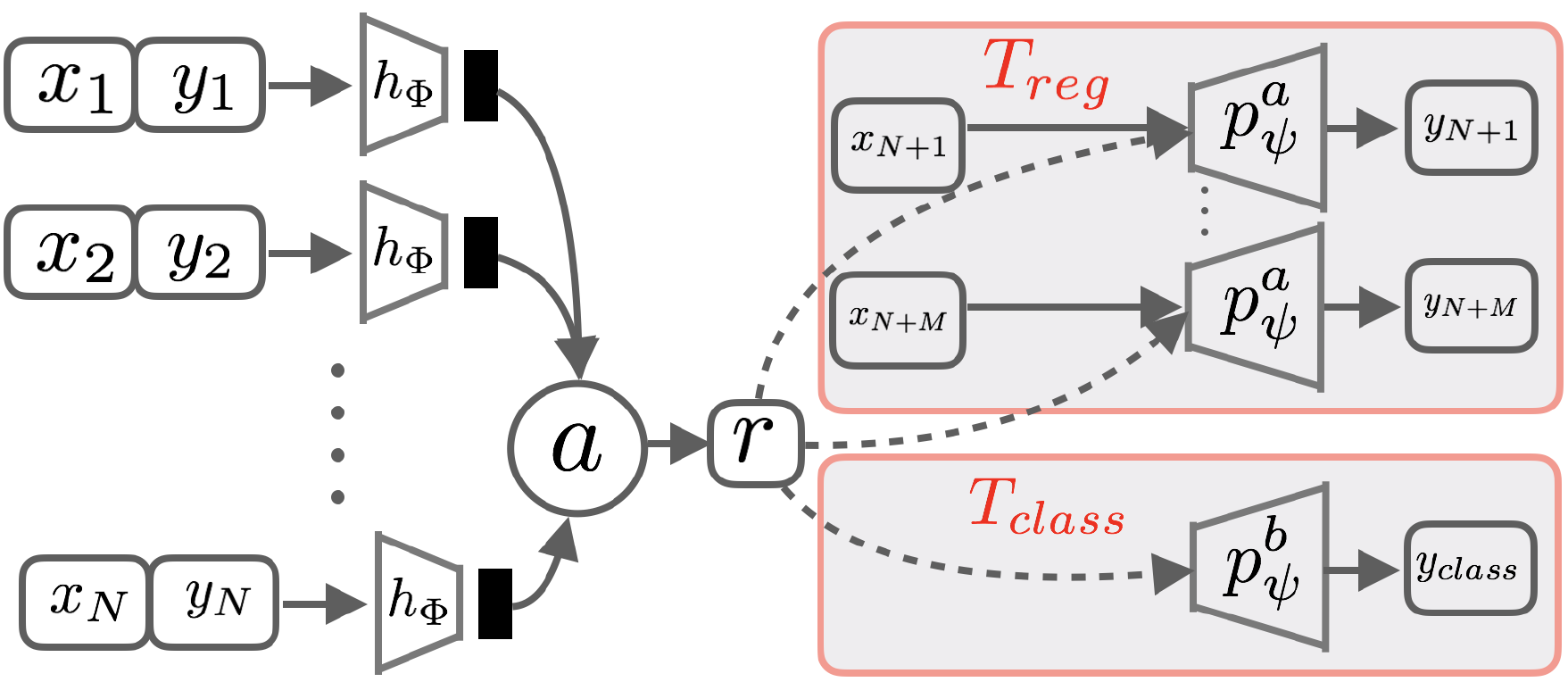}
\caption{FCRL (transfer)}
\end{subfigure}
\vspace{-2mm}
\caption{
The difference in the training of CNP \citep{garnelo2018conditional} and FCRL for learning meta-representations $r$ of functions. (shown left) CNP learns the aggregated representation $r$ of the context set by maximizing the conditional likelihood of the target data. (shown center) Training of FCRL encoder $h_{\phi}$ by contrasting context sets of different functions. Note that the target inputs $x_{N+1}$ etc., are not used at this stage. (shown right) Using the pretrained FCRL encoder $h_{\phi}$, we train separate decoders $p_{\psi}^{*}$ for each downstream task, shown in grey boxes. The dotted arrows indicate the transfer of inferred meta-representations to the tasks.
}\label{figure:comparison}
\vspace{-1mm}
\end{figure*}

However, CNPs optimize encoder jointly with the decoder on the downstream prediction task, i.e., prediction of function values $y$ at unseen locations $x$, as illustrated in \Cref{figure:comparison}(a). This ties the meta-representation's quality 
to the combined encoder and decoder performances on this particular task and thereby makes it susceptible to supervision collapse, i.e. the representations lose any information which is irrelevant for solving the training task, but may be necessary for the transfer to new tasks \citep{doersch2020crosstransformers}. Moreover, many real-world tasks are noisy, and the prediction task might entail reconstructing high dimensional data, such as images in GQNs \citep{eslami2018neural}. The corresponding objective function can cause the model to waste its capacity on reconstructing unimportant features such as static backgrounds and noise, while ignoring visually small but important details in its learned representation \citep{anand2019unsupervised, kipf2019contrastive}. This is crucial for many real-world applications; for instance, in order to manipulate a small object in a complex scene, the model's ability to infer the object's shape carries more importance than inferring its color or reconstructing the static background.

In this work, we study the generalization of a function's meta-representations in terms of their transferability to downstream tasks and their robustness to noise. We empirically show that the joint optimization of meta-representations and a prediction task is detrimental to the transferability of meta-representations and makes them vulnerable to noise. To address this issue, we propose a decoupled encoder-decoder training scheme, wherein the encoder is exclusively trained by a novel contrastive learning framework which we call FCRL (Function Contrastive Representation Learning). Instead of relying on reconstructions, it learns by contrasting sets of input-output pairs sampled from different functions. The key idea is that two sets of samples from the same function should have similar latent representations, while representations of sets of samples from different functions should remain easily distinguishable. FCRL retains the useful properties of meta-representations such as shared learning and sample efficiency while improving its transferability to downstream tasks and robustness to noise. Unlike contemporary meta-learning algorithms, meta-representations in FCRL are explicitly optimized over a distribution of functions rather than tasks.

To evaluate the effectiveness of the proposed method, we conduct comprehensive experiments on diverse downstream problems, including classification, regression, parameter identification, scene understanding, scene reconstruction and reinforcement learning. We consider different datasets, ranging from simple 1D and 2D regression to challenging simulated and real-world scenes. In particular, we find that a downstream predictor trained with our (pre-trained) encoder compares favorably to related methods on these tasks, including ones where the predictor is trained jointly with the encoder.

\vspace*{-1mm}
\section{Preliminaries}
\label{background}
\vspace*{-1mm}
\subsection{Problem Setting}
Consider a distribution over data-generating functions $p(f)$. Let $f$ be a sample from this distribution $f \sim p(f)$, where $f: \mathcal{X} \to \mathcal{Y} $ with $\mathcal{X} = \mathbb{R}^d$ and $\mathcal{Y} \subseteq \mathbb{R}^{d'}$:
\begin{equation}\label{eq:func}
y = f(x, \xi); \; ~~\xi \sim \mathcal{Z}
\end{equation}
where $\xi$ is sampled from some noise distribution $\mathcal{Z}$. 
Let $O^f =\{(x_i, y_i)\}_{i=1}^N$ be a set of few observed examples of a function $f$, referred to as the context set, and consider a set of downstream tasks $\mathcal{T}$. Here, each task $T \in \mathcal{T}$ can be defined as a mapping defined over $f$. In the case of few shot regression (see Section~\ref{sec:1d-2d-regression}), $T$ maps from $f$ to a predictive model $p_{\psi}(y | x)$. In the case of parameter identification, $T$ maps from $f$ to some scalar or vector valued parameter of $f$.
Our goal is therefore to learn an encoder which maps a context set $O^f$ to a representation of $f$ that
can interchangeably be used for multiple downstream tasks $\mathcal{T}$ defined on the same function (without requiring retraining).

\subsection{Background}
In this section, we briefly discuss a class of meta-learning methods that are particularly relevant to our encoder-decoder setting, namely conditional neural processes (CNPs) and generative query networks (GQNs) \citep{garnelo2018conditional, garnelo2018neural, eslami2018neural}.

\textbf{Conditional Neural Processes (CNPs).} The key proposal in CNPs (applied to few-shot learning) is to express a distribution over predictor functions given a context set. They learn the meta-representations $r$ by jointly training the encoder and decoder, as illustrated in \Cref{figure:comparison}(a). To this end, they first encode the context $O^{f}$ into individual representations $r_i = h_{\Phi}(x_i, y_i)~ \forall  i \in [N]$, where $h_{\Phi}$ is a neural network. The representations are then aggregated via a mean-pooling operation into a fixed size vector $r = \nicefrac{1}{N} (r_1 + r_2 + ... +  r_N)$. The idea is that $r$ captures all the relevant information about $f$ from the context set $O^{f}$; accordingly, the predictive distribution is approximated by maximizing the conditional likelihood of the target distribution $p(y|x, O^f)$, where $y = f(x)$.

\textbf{Generative Query Networks (GQN).} GQN \citep{eslami2018neural} can be seen as an extension of NPs \citep{garnelo2018neural} for learning 3D scenes representations. The context dataset $O^{f}$ in GQN consists of tuples of camera viewpoints in 3D space ($\mathcal{X}$) and the images taken from those viewpoints ($\mathcal{Y}$). Like NPs, GQNs learn to infer the latent representation of the scene (a function) by conditioning on the aggregated context and maximizing the likelihood of generating the correct image corresponding to a queried viewpoint.

\section{Function-Contrastive Representation Learning (FCRL)}\label{sec:method}
We take the perspective here that the sets of context points $O^f$ provide a partial observation of an underlying function $f$. Our goal is to find an encoder $g_{(\phi, \Phi)}$ which maps such partial observations to low-dimensional representations of the underlying function. The key idea is that a good encoder $g_{(\phi, \Phi)}$ should map different context sets (i.e. partial observations) of the same function to be close in the latent space, such that they can easily be identified among context sets of different functions. This motivates the contrastive-learning objective which we will detail in the following.

\textbf{Encoder Structure.} Since the inputs to the encoder $g_{(\phi, \Phi)}$ are sets, it needs to
be permutation invariant with respect to input order and able to process inputs of varying sizes. We enforce this permutation invariance in $g_{(\phi, \Phi)}$ via sum-decomposition, proposed by \citep{zaheer2017deep}. We first average-pool the point-wise transformations of $O^f$ to get the encoded representations
\begin{equation}\label{eq:11}
r^f = \frac{1}{|O^f|}\sum_{(x,y)\in O^f}h_{\Phi}(x, y)
\end{equation}
where $h_{\Phi}(\cdot)$ is the encoder network. We then obtain a nonlinear projection of this encoded representation $g_{(\phi, \Phi)}(O^f) = \rho_{\phi}(r^f)$. Note that the function $\rho_{\phi}$ can be any nonlinear function. We use an MLP with one hidden layer which also acts as the projection head for learning the representation. Similar to \citep{chen2020simple}, we found that it is beneficial to define the contrastive objective on these projected representations $\rho_{\phi}(r^f)$, rather than directly on the encoded representations $r^f$. More details can be found in our ablation study in \Cref{sec:ablations_appendix}.

\begin{figure}%
	\centering
	\includegraphics[width=0.49\textwidth]{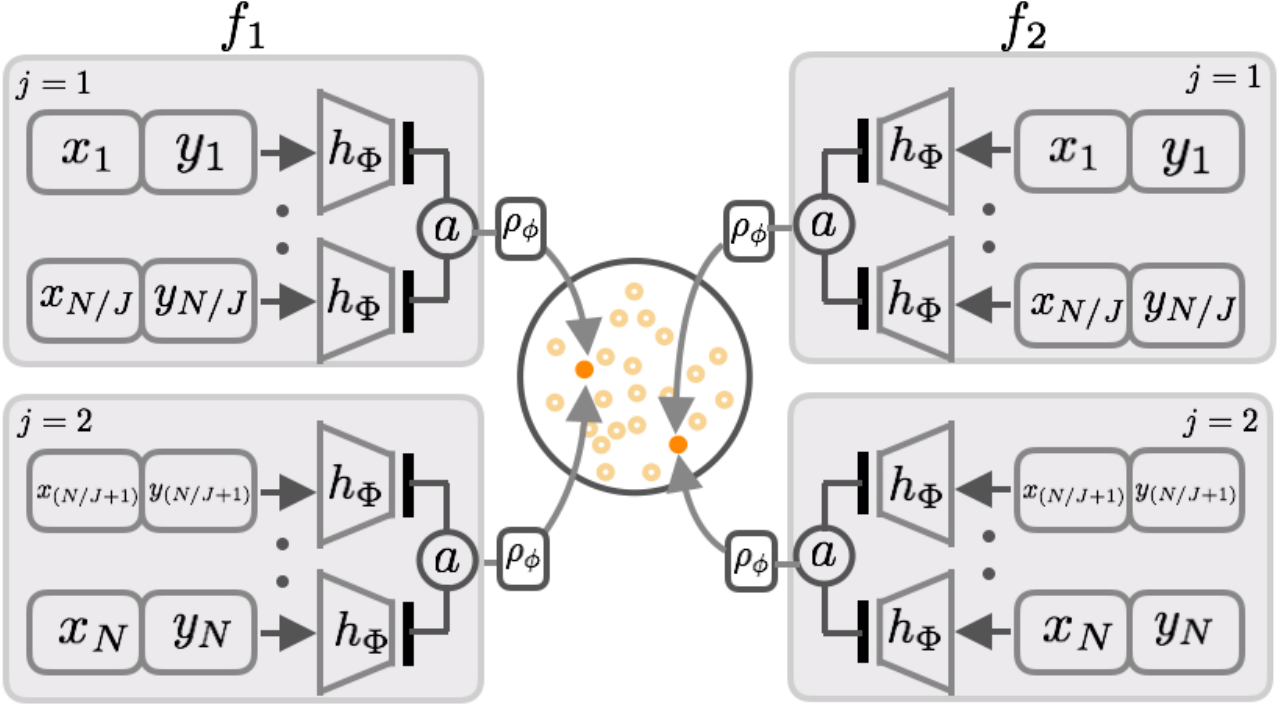}
	\vspace{-4mm}
	\caption{Inner-workings of the FCRL objective function. We split the context set of each function into $J$ disjoint views, and align the aggregated representations of those views. Shown here is the example of two functions, with two views each i.e., $J=2$.}
	\vspace{-2mm}
	\label{fig:loss}
\end{figure}%

\textbf{Encoder Training.} At training time, we are provided with  partial observations $O^{1:K}$ from $K$ functions. Each observation is a set of $N$ i.i.d.\ (independent and identically distributed) samples $O^{k}=\{(x_i^k, y_i^k)\}_{i=1}^{N}$. To encourage that different observations of the same functions are mapped to similar representations, we will now formulate a contrastive learning objective, as illustrated in \Cref{fig:loss}. To apply contrastive learning, we create different views of the same function $k$ by splitting each sample set $O^{k}$ into $J$ subsets of size $N/J$. Defining $t_{j} := \{(j-1)N/J+1,...,jN/J\}$, we obtain a split of $O^{k}$ into $J$ disjoint subsets of equal size:
\begin{equation}\label{eq:1}
O^{k}=\cup_{j=1}^{J}O_{t_{j}}^{k} \; \text{with} \; O_{t_{i}}^{k} \cap O_{t_{j}}^{k} = \emptyset \, \text{if} \, i \ne j
\end{equation}
where each subset $O_{t_{j}}^{k}$ is a partial view of the underlying function $k$.
Here, $J$ is a hyper-parameter and can vary in the range of $[2, N]$ and must divide $N$, i.e. $N \mod J = 0$. 
Its value is empirically chosen based on the data domain: for 1D and 2D regression problems (Sections~\ref{sec:1d-2d-regression}), the number of examples per view ($N/J$) is relatively large as a single context point does not provide much information about the underlying function; whereas in scenes (Section~\ref{sec:scenes}), a few (or even one) images per view (partial observation) may provide enough information. We expand on the appropriate choice of $J$ in our experiments and the respective ablations. For brevity of notation, we define $v_j^k:=g_{(\phi, \Phi)}(O_{t_{j}}^{k})$. We now formulate the contrastive learning objective as follows:
\begin{equation}\label{eq:2}
\hspace{-1mm}\mathcal{L}\hspace{-1mm}=\hspace{-1mm}\sum_{k=1}^{K}\sum_{1\le i<j\le J}\left[\log\frac{\exp\left(\mathrm{sim}\left(v_j^k, v_i^k\right) / \tau \right)}{\sum_{m=1}^{K}\exp\left(\mathrm{sim}\left(v_j^k, v_i^m\right)/ \tau \right)}\right]\hspace{-1mm}
\end{equation}
where $\mathrm{sim}(a,b):=\frac{a^\top b}{\|a\| \|b\|}$ is the cosine similarity measure. 
Intuitively, the objective function in \Cref{eq:2} encourages that the similarity measure $\mathrm{sim}(v_{(\cdot)}^p, v_{(\cdot)}^q)$ acts as a discriminatory function, yielding a large value if $v_{(\cdot)}^p$ and $v_{(\cdot)}^q$ are representations of sets of samples drawn from the same function, i.e. if $p = q$ (\emph{positives}), and a small value otherwise, i.e. if $p \ne q$ (\emph{negatives}). 
The second summation in \Cref{eq:2} over $1 \le i \le j \le J$ is over available pairs of positives, and $\tau$ is a temperature parameter which scales the scores returned by the similarity measure. Similar to SimCLR \citep{chen2020simple}, we find that temperature adjustment is important for learning good representations and treat it as a hyperparameter (ablation study in \Cref{sec:ablations_appendix}).

We note that the learning objective effectively balances two goals. The first is that of avoiding overfitting (i.e., regularization). It encourages that any two independent samples from the same distribution get mapped to similar points. This is akin to the method of ``symmetrization by a ghost sample'' which is a standard trick in proving learning theory bounds \citep{Vapnik95}. Essentially, if two means on different samples are close, then they will also be close to their expectation, i.e., they will not overfit to the data.\footnote{
This is an example of the more general phenomenon of concentration of measure, applicable not just to means but also to other functions that aggregate samples. For a simple argument, let $\mathbb{E}_{O^i}$ denote the expectation w.r.t.\ drawing the sample $O^i$, and $g$ be the mapping function applied to two independent samples $O^1, O^2$ from the same distribution. Then we have 
$| g(O^1) - \mathbb{E}_{O^1}[ g(O^1) ] |
=
| g(O^1) - \mathbb{E}_{O^2}[ g(O^2) ] |
=
| \mathbb{E}_{O^2}[  g(O^1) - g(O^2)  ] |
\le
\mathbb{E}_{O^2}[ |  g(O^1) - g(O^2) | ]
$. The second equality uses independence of the samples $O^1$ and $O^2$, and the last step uses Jensen's inequality. This shows that if in expectation the embeddings of two samples are close (r.h.s.), then each embedding is close to its expectation. \vspace{-3mm}} In spirit, this is close also to the idea of regularization by enforcing stability (i.e., weak dependence on sampling points) \citep{Bousquet02}.
The second goal is to preserve contrastive information, ensuring that samples from different distributions get mapped to different points. Both goals are intricately linked in our setting, where the aggregation function is being learnt, since the second component is necessary to prevent the system from trivially meeting the first goal by, say, mapping everything to 0. 

\vspace{-1mm}
\subsection{Application to Downstream Tasks}
\vspace{-1mm}
Once representation learning using FCRL is concluded, $h_{\Phi}$ is fixed and can now be used for
few-shot downstream prediction tasks $\mathcal{T}$ defined on the underlying data-generating functions. To solve a particular downstream task, one may optimize a parametric decoder $p_{\psi}(\cdot|r)$ conditioned on the learned representation $r$. Specifically, the decoder maps the representations learned in the previous step to the variable of interest in the given task. Depending on the nature of the downstream task, the conditional distributions and the associated objectives can be defined in different ways.

\vspace{-1mm}
\section{Experiments}
\vspace{-1mm}
To illustrate the benefits of learning function representations without an explicit decoder, we consider four different experimental settings. In all experiments that follow, we first learn the encoder, and then keep it fixed. Subsequently, we optimize decoders for the specific downstream problems at hand, while keeping the meta-representations from the encoder detached from the computational graph.

\textbf{Baselines.} We compare the downstream predictive performance of FCRL based representations with the representations learned by the closest task-oriented, meta-learning baselines. For a fair comparison, all the baselines and FCRL have the same encoding architecture. For instance, for 1D and 2D regression functions, we consider CNPs and NPs as baselines. 
We share with these methods an identical way of mapping the context set to its representation, but unlike us, they optimize directly for the performance of the decoder $p(y | x)$ jointly with the said representation. 
For scene datasets, we use GQN (a variant of NPs) as the baseline, one that explicitly learns to reconstruct scenes using a limited number of context samples (pairs of camera viewpoints and the corresponding images).
\vspace{-1mm}
\subsection{1D and 2D Functions}
\label{sec:1d-2d-regression}
\vspace{-1mm}
In the first set of experiments, we consider two different distributions of functions i.e., a distribution over 1D sinusoidal waves, proposed by \citep{finn2017model}, and a relatively harder distribution where images are modelled as 2D functions \citep{garnelo2018conditional,garnelo2018neural,gordon2020convolutional, kim2019attentive}. The representation learning stage for both datasets is similar, however the downstream tasks differ.

\textbf{1D Sinusoidal Functions:} We consider a dataset of $20,000$ training, $1000$ validation and $1000$ test sinusoidal functions. The amplitude and the phase of the functions are sampled uniformly from $[0.1, 0.5]$ and $[0, \pi]$ respectively. For each function $f$, the $x$-coordinates are uniformly sampled from $[-5.0, 5.0]$ and then $f$ is applied to obtain the $y$-coordinates. 

\textbf{Modeling Images as 2D Functions:} In this setting, each image is regarded as a function mapping from 2D pixel coordinates (comprising function input $x_i$) to the pixel intensities at the corresponding pixel coordinate (comprising function output $y_i$). We consider images of MNIST digits \citep{lecun1998gradient}, where $x_i \in [0,1]^2$ are the normalized pixel coordinates and $y_i \in [0,1]$ is a grayscale pixel intensity. The training and validation datasets consists of $60,000$ MNIST training and $10,000$ test samples, respectively.
\vspace{-1mm}
\subsubsection{Representation Learning Stage}
\vspace{-1mm}
We first describe the representation learning stage for both datasets, and then provide results on their respective downstream tasks. For training the encoder $g_{(\phi, \Phi)}$, we have a dataset $O=\{O^k\}_{k=1}^K$ at our disposition, where each $k$ corresponds to a function $f_k$ which has been sampled as described above. Each individual sample $O^k=\{(x_i^k,y_i^k)\}_{i=1}^{N}$ from the dataset is itself a set, comprising $N$ input-output pairs from that particular function $f_k$, i.e. $y_i^k = f_k(x_i^k)$. For sinusoidal functions, we fix the maximum number of context points to 20 and the number of examples $N$ is chosen randomly in $[2,20]$ for each $k$. For MNIST digits as 2D functions, we allow a maximum of 200 samples per context set, and $N$ is sampled randomly from $[2, 200]$ for each $k$. The encoder $g_{(\phi, \Phi)}$ is then trained by splitting each context set $O^k$ into $J$ disjoint views. We set $J=2$ for the sinusoidal functions and $J=10$ for the 2D functions. An ablation study for the choice of $J$ is presented in \Cref{sec:ablations_appendix}. 

\subsubsection{Downstream Tasks on 1D Sinusoids}
\label{sec:1d-tasks}
After training the encoder $g_{(\phi, \Phi)}$, we discard the projection head $\rho_{\phi}$ and use the trained encoder $h_{\Phi}$ to extract the representations. For 1D sinusoids, we define two downstream tasks on the learned representation: $\mathcal{T}_{1D} = \{T_{fsr}, T_{fspi}\}$, where $T_{fsr}$ and $T_{fspi}$ are few-shot regression and few-shot parameter identification tasks, respectively. The decoders for the downstream tasks are trained as follows:

\begin{figure}%
	\centering
	\includegraphics[width=0.48\textwidth]{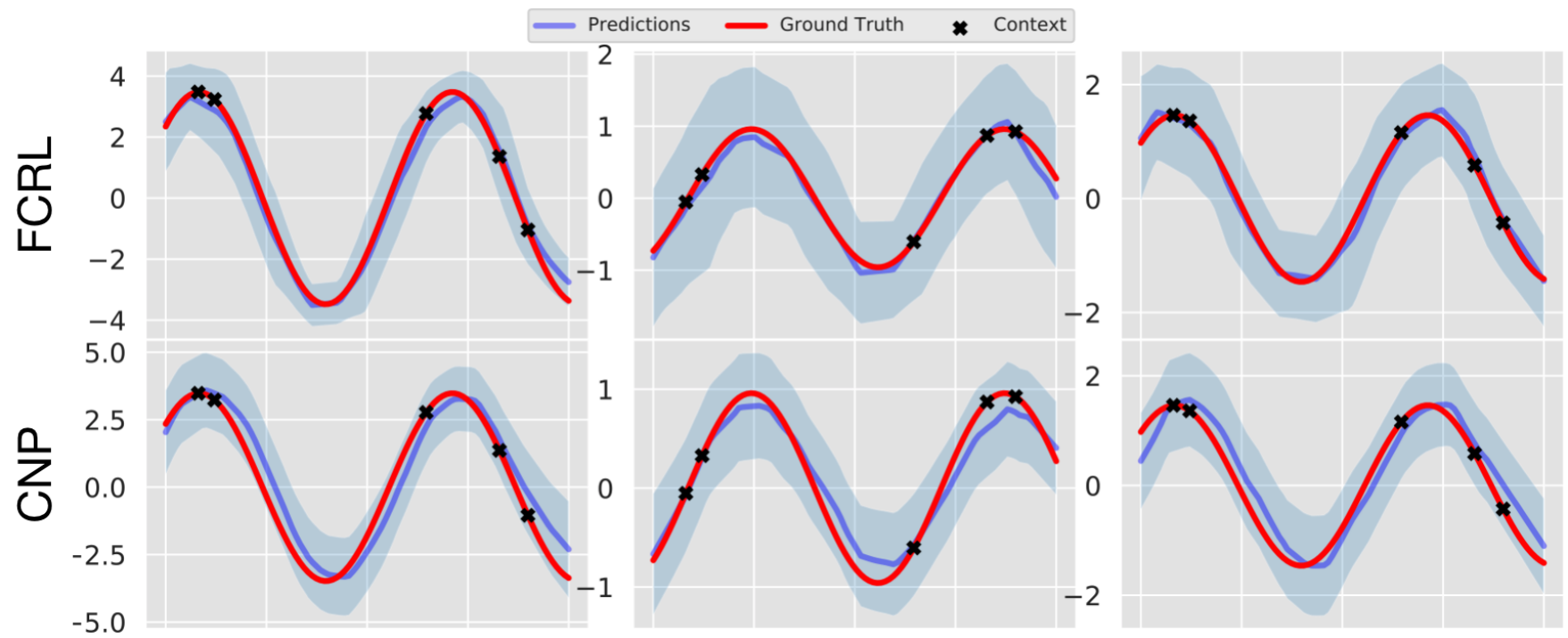}
	\vspace{-4mm}
	\caption{Qualitative comparison of different methods on $5$-shot regression task on three different sinusoids. The decoder trained with FCRL representation predicts the correct form of the sinusoids.}
	\label{fig:sinusoid}
	\vspace{-2mm}
\end{figure}%

\textbf{Few-shot Regression (FSR).} FSR for 1D functions is a well-studied problem in meta-learning \citep{garnelo2018conditional,finn2017model, kim2019attentive,xu2019metafun}. For each sampled function $f_k$, we are given a context set $O^{k} = \{(x_i^k, y_i^k = f_k(x_i^k))\}_{i = 1}^{N}$ of size $N$, which can be utilized to infer the meta-representation $r^k$ of $f_k$ via the pre-trained encoder $h_{\Phi}$. We are then provided with $M$ additional samples from $f_k$ (not seen by the encoder $h_{\Phi}$). The goal for a downstream decoder is to predict $y_i^k$, given $x_i^k$ and the meta-representation $r^k$. In other words, the downstream decoder with parameters $\psi$ models the distribution $p_{\psi}(y_i^k | x_i^k, r^k)$. Where $D^k = \{(x_i^k, y_i^k)\}_{i=1}^{N + M}$, the decoder is therefore trained to solve the following problem:
\begin{equation}\label{eq:12}
    \max_{\psi}\mathop{\mathbb{E}_{f_{k}\sim p(f)}}\left[\mathop{\mathbb{E}}_{(x_i^k, y_i^k) \sim D^k}[\log p_{\psi}(y^k_i |x^k_i,r^k)]\right]
\end{equation}
Here, the value of $M$ is sampled randomly from the interval $[0, 20 - N]$. The decoder $p_{\psi}$ is an MLP with two hidden layers and it is trained with the same training functions as the encoder $h_{\Phi}$. 
In addition to the Gaussian mean of $y_i^k$, it also outputs the variance in order to quantify the uncertainty in the point estimates. The qualitative results on test functions as shown in \Cref{fig:sinusoid} demonstrate that our model is able to quickly adapt with as few as $5$ context points. In \Cref{tab:sinusoid}, we compare our method with CNP and NP quantitatively, and show that the predictions of our method are closer to the groundtruth, even though the encoder and decoder in both CNP and NP are explicitly trained to directly maximize the log likelihood to fit the sinusoid.

\textbf{Few-shot Parameter Identification (FSPI).} The goal here is to predict the amplitude ($y^k_{amp}$) and phase ($y^k_{phase}$) of the sampled sine wave $f_k$, given a context set $O^{k} = \{(x_i^k, y_i^k = f_k(x_i^k))\}_{i = 1}^{N}$ of $N$ samples. Having encoded the context set $O^k$ to meta-representation $r^k$ via the pre-trained encoder $h_{\Phi}$ (following \Cref{eq:11}), we train a linear decoder $p_{\psi}$ on top of the said representation by maximizing the likelihood of the sine wave parameters. The predictive distribution is $p_{\psi}(y^{k}_{amp}, y^{k}_{phase}|r^k)$ and the objective is:
\begin{equation}\label{eq:13}
    \max_{\psi}\mathop{\mathbb{E}_{f_k\sim p(f)}}[\log p_{\psi}(y^k_{amp}, y^k_{phase}|r^k)]
\end{equation}
Similar to FSR, we use the same training functions to train $p_{\psi}$ as we did to train the encoder $h_{\Phi}$. In \Cref{tab:sinusoid}, we report the mean squared error for three independent runs, averaged across all the test tasks for $5$-shots and $20$-shots FSR and FSPI. In both prediction tasks, the decoders trained on FCRL representations outperform CNP and NP. More details on the experiment are given in \Cref{sec:exp-1D}.

\begin{table}%
    \centering
    \begin{tabular}{l|cc}
    \toprule
    & \multicolumn{2}{c}{\textbf{Few-shot Regression (FSR)}} \\
    \cmidrule(r){2-3}
    Models    & \textbf{5-shot}     & \textbf{20-shot} \\
    \midrule
    NP  & $0.310\pm0.05$ & $0.218\pm0.02$ \\
    CNP  & $0.265\pm0.03$ & $0.149\pm0.02$ \\
    FCRL & $\bm{0.172\pm0.04}$ & $\bm{0.100\pm0.02}$  \\
    \end{tabular}
    
    \begin{tabular}{l|cc}
    \toprule
    & \multicolumn{2}{c}{\textbf{Few-shot Parameter Identification}} \\
    \cmidrule(r){2-3}
    Models    & \textbf{5-shot}     & \textbf{20-shot} \\
    \midrule
    NP  & $0.0087\pm0.0007$ & $0.0037\pm0.0005$ \\
    CNP  & $0.0096\pm0.0007$ & $0.0049\pm0.0011$\\
    FCRL & $\bm{0.0078\pm0.0004}$ & $\bm{0.0032\pm0.0002}$  \\
    \bottomrule
    \end{tabular}
    \caption{Mean squared error (MSE) for all the target points in $5$ and $20$ shot regression and parameter identification tasks on test sinusoid functions. The reported values are the mean and standard deviation of three independent runs. FCRL performs slightly better than CNP and NP on both tasks.}
    \label{tab:sinusoid}
\end{table}

\subsubsection{Downstream Tasks on 2D Functions}
\label{sec:2d-tasks}
Similar to the tasks above, after training the model $g_{(\phi, \Phi)}$, we discard the projection head $\rho_{\phi}$ and use the trained encoder $h_{\Phi}$ to extract the representations. For MNIST digits as functions, we formulate two downstream prediction tasks on the learned representations: $\mathcal{T}_{2D} = \{T_{fsic}, T_{fscc}\}$, where $T_{fsci}$ corresponds to few-shot image completion and $T_{fscc}$ corresponds to few-shot content classification task. The decoders for the downstream tasks are trained as following.

\begin{figure}[h!]
	\centering
	\includegraphics[width=0.46\textwidth]{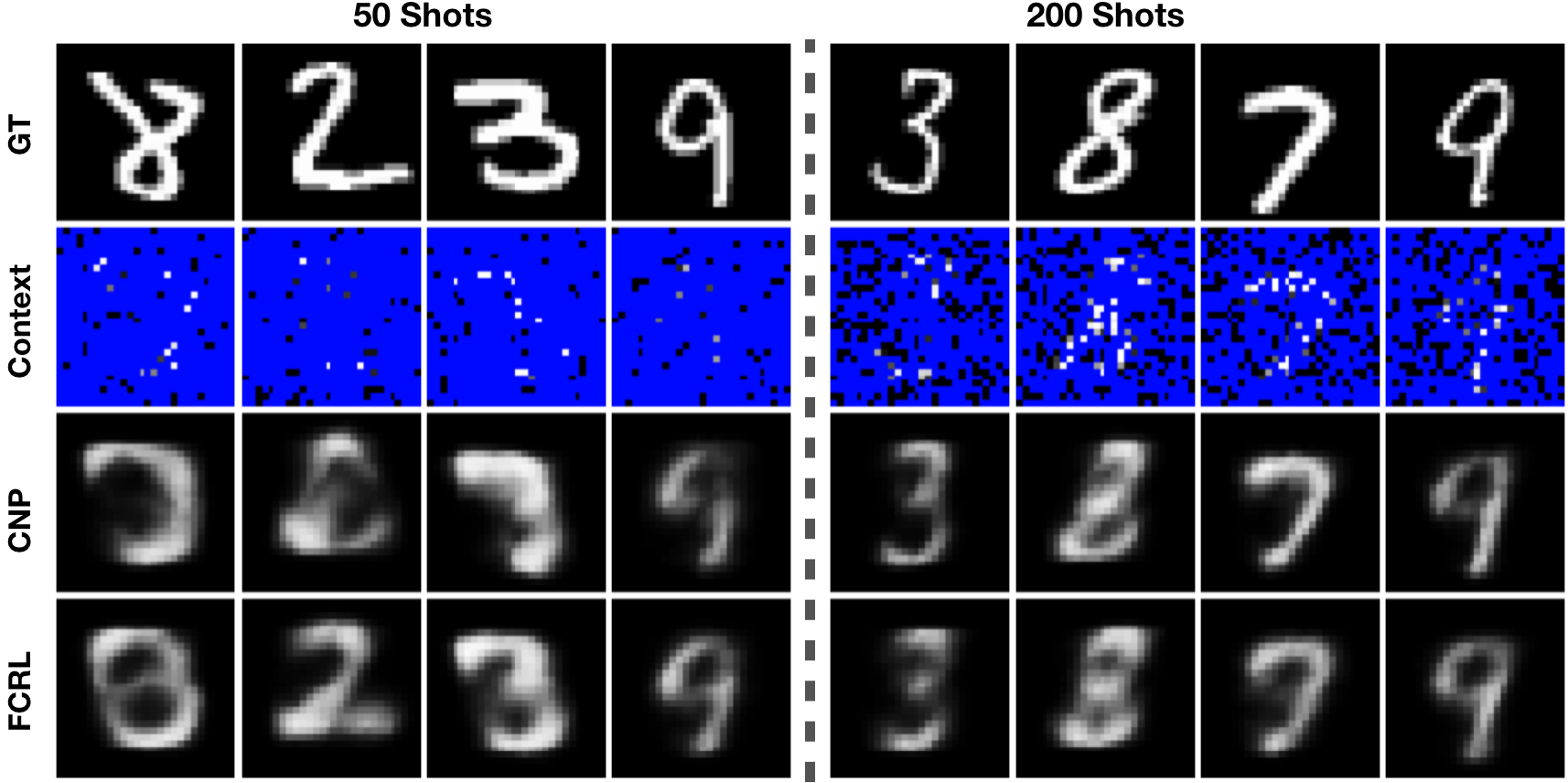}
	\caption{Qualitative comparison of CNP and FCRL based few-shot image completion. Here, each digit corresponds to one function. The context is shown in the second row where the target pixels are colored blue. FCRL is comparable with CNPs at predicting the correct form of a digit.}
	\label{fig:mnist_qualitative}
\end{figure}

\textbf{Few-shot Content Classification (FSCC).}
To evaluate how much semantic information is captured by the meta-representations, we propose the task of few-shot content classification (FSCC). The goal here is to predict the class of each MNIST image given a context set $O^k = \{(x_i^k,y_i^k)\}_{i=1}^{N}$ comprising a few randomly sampled pixel coordinates $x_i^k$ and the corresponding grayscale intensities $y_i^k$. The lack of explicit spatial structure in the context points makes it a challenging problem.
\begin{figure}%
\begin{subfigure}{0.235\textwidth}
\centering\includegraphics[width=\textwidth]{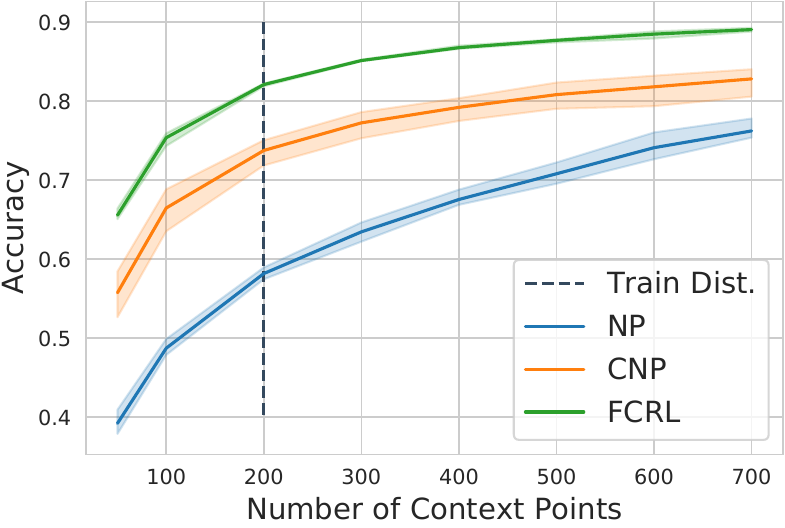}
\end{subfigure}\hspace{0.5mm}
\begin{subfigure}{0.235\textwidth}
\centering\includegraphics[width=\textwidth]{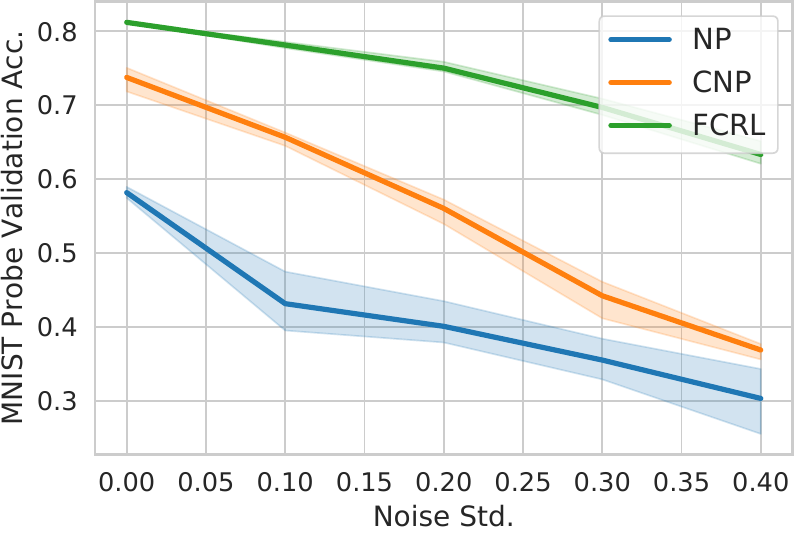}
\end{subfigure} %
\caption{(left) Quantitative evaluation of the models in terms of digit classification from the fixed number of context points (varying along the x-axis). The error bands show the standard deviation over three runs. FCRL achieves substantially higher accuracy than both baselines for all evaluated numbers of context points.(right) Quantitative comparison for robustness to noise on MNIST content classification downstream task. The representations learned with FCRL are much more robust to noise than with CNPs and NPs.}
\label{figure:mnist_probe_noise}
\vspace{-3mm}
\end{figure}
We use the pre-trained encoder $h_{\Phi}$ to encode $O^k$ to its representation $r^k$, and train a linear decoder on top to classify the class label $y^k_{one\_hot}$ corresponding to the MNIST image from which $O^k$ is sampled. The decoder $p_{\psi}$ therefore solves the following classification problem:
\begin{equation}\label{eq:14}
\vspace{-1mm}
    \max_{\psi}\mathop{\mathbb{E}_{f_k\sim p(f)}}[\log p_{\psi}(y^k_{one\_hot}|r^k)]
\vspace{-1mm}
\end{equation}
We train the decoder with the same functions (images) that were used for training the encoder $h_{\Phi}$, and subsequently evaluate them on unseen functions from the validation set. \Cref{fig:mnist_qualitative} shows the performance of decoders applied to representations obtained from FCRL, CNP and NP for varying size of the context set $O^k$. We find that FCRL significantly outperforms the baselines at any given number of context points, suggesting that the encoder $h_{\Phi}$ is able to efficiently extract semantic information in an unsupervised manner. We also observe that it is able to generalize to larger number of context points than encountered during training ($200$).

\begin{figure*}%
	\centering
	\includegraphics[width=\textwidth]{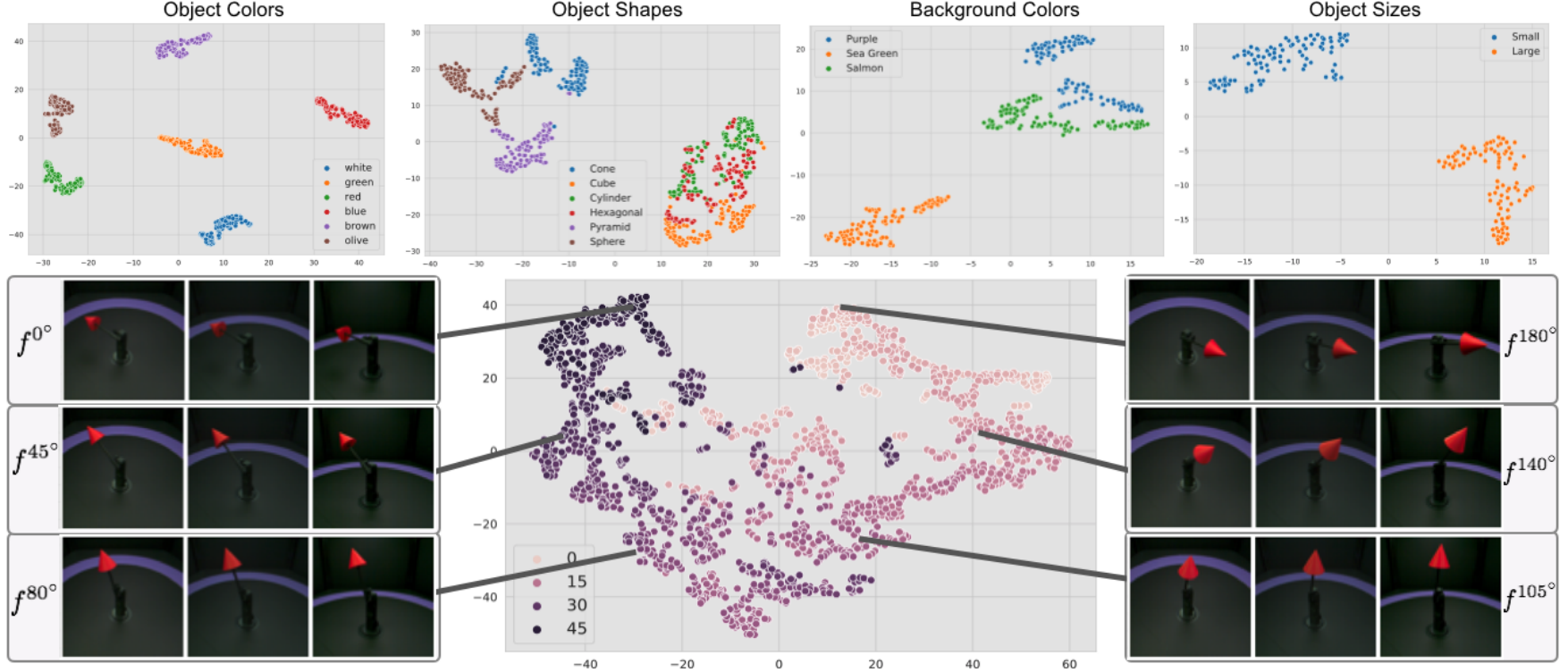}
	\caption{tSNE projections of the meta-representations learned by FCRL on MPI3D dataset (treating scene as functions). (top row) Each plot shows the latent structure corresponding to the factors mentioned. (bottom row) Shows the latent structure corresponding to the factor of robot arm rotation along $1^{st}$ DOF. Each embedding corresponds to the aggregated representation of three views of the same scene, denoted as $f^{angle^{\circ}}$. It can be seen that by learning to distinguish between functions, FCRL captures the semantic underlying structure of the functions' distribution. \emph{Note:} Each plot is generated by varying the factor of interest and keeping the rest of the factors fixed, except the factors of the first degree of freedom and the second degree of freedom.}
	\label{fig:mpi3d_latents}
	\vspace{-5mm}
\end{figure*}%

\textbf{Few-shot Image Completion (FSIC).}
This setting is identical to that of Few-Shot Regression (FSR) described in \Cref{sec:1d-tasks}, except $M$ is sampled randomly from $[0, 200-N]$ and the decoder is a two-layer MLP with two input units (to account for the fact that the input $x_i^k$ is now two dimensional). Qualitative results of FSIC on test images are shown in \Cref{fig:mnist_qualitative}. It can be seen that the decoder trained on FCRL representations is able to predict the pixel intensities reasonably well, even when the number of context points is as low as 50, or approximately $6\%$ of the image. We compare its performance against CNP, which uses the same parameterization of both the encoder and the decoder. We however note a crucial distinction: in FCRL, the meta-representation (resulting from the encoder) is not optimized for the image completion task. In particular, no gradient flows from the decoder to the encoder, and the former is trained independently of the latter. On the contrary, CNP jointly optimizes both encoder and decoder parameters to solve the image completion task (i.e. to predict the pixel values). Despite the fact, it appears that the quality of reconstructions from the FCRL decoder matches that from the CNP decoder.

We note that the gap between CNP and FCRL is reduced if the training scheme aligns with the downstream task. In FSIC, the downstream task is to obtain a generative model which is exactly what CNPs are trained for, therefore CNPs tend to perform on par with FCRL as shown in \Cref{tab:fsic-mse}.

\begin{table}%
    \centering
    \begin{tabular}{l|cc}
    \toprule
    & \multicolumn{2}{c}{\textbf{Few-shot Image Completion (FSIC)}} \\
    \cmidrule(r){2-3}
    Models    & \textbf{50-shot}     & \textbf{200-shot} \\
    \midrule
    NP  & $0.0531\pm\num{0.0002}$ & $0.0424\pm\num{0.0002}$ \\
    CNP  & $0.0477\pm\num{0.0006}$ & $0.0347\pm\num{0.0011}$ \\
    FCRL & $0.0481\pm\num{0.0001}$ & $0.0355\pm\num{0.0001}$  \\
    \bottomrule
    \end{tabular}
    \caption{MSE for all the target points in $50$ and $200$ shot image completion task on MNIST as functions. The reported values are the mean and standard deviation of three independent runs.}
    \label{tab:fsic-mse}
\end{table}

\subsection{Robustness to Noise Corruptions}
In our experiments so far, we have considered the functions to be deterministic. However in real-world settings, data-generating functions are corrupted with noise. 
In this section, we assume that they take the form:
\begin{equation}
    y = f(x, \xi) = f(x) + \xi; \; \text{where} \; \xi \in \mathcal{N}(0, \sigma)
\end{equation}
where $\mathcal{N}(0, \sigma)$ is the standard Gaussian distribution with standard deviation $\sigma$. We now investigate the robustness of FCRL and the baselines as $\sigma$ is varied. To this end,
we train all the models on the noisy data and evaluate the quality of the learned representation on the Few-Shot Content Classification downstream task, as defined above. We find that the representations learned by FCRL to be significantly more robust to increasing noise strength ($\sigma$) than the baselines, as illustrated in \Cref{figure:mnist_probe_noise}.
One possible explanation for the susceptibility of CNP and NP to noise is the fact that the representations are learned by reconstructing the outputs, where signal to noise ratio is low. On the other hand, FCRL learns by contrasting the set of examples, extracting only the invariant features and discarding non-correlated noise in the input. Similar results on scene understanding datasets are presented in \Cref{sec:noise_robustness_appendix}.

\subsection{Representing Scenes as Functions} \label{sec:scenes}
Like \citet{eslami2018neural}, we represent scenes as deterministic functions which map camera viewpoints to images. 
Precisely, each such scene is represented by a function $f_k$, and we consider context sets $O^k=\{(x_i^k,y_i^k)\}_{i=1}^{N}$, where $x_i$ is the 3D camera viewpoint and $y_i$ is the corresponding image taken from that viewpoint. Given this set of viewpoint-image pairs, we apply the proposed method on $O^k$ to obtain a representation of the scene, $r^k$. The usefulness of this representation is then evaluated for three downstream tasks: scene understanding, scene reconstruction and reinforcement learning (RL). 

For the first task, our goal is to determine whether the representation $r^k$ contains enough information to infer the underlying factors of variation \citep{bengio2013representation} of a given scene. To this end, we use MPI3D \citep{gondal2019transfer}, a real-world robotics dataset comprising pairs of images from three camera viewpoints and the corresponding factors of variation (including the position, orientation, size and color of an object in the scene). For the second task, we analyze whether the learned representation $r^k$ can be used to reconstruct the scene from an unseen viewpoint. This objective is similar to what GQNs \citep{eslami2018neural} are originally trained for, serving as an ideal baseline for this task. For the last task, our objective is to determine whether $r^k$ contains enough useful information to guide a RL agent towards maximizing its reward. To this end, we create RLScenes, a multi-view robotics dataset based on an open-source physics simulation engine (details in \Cref{sec:scenes_datasets_appendix}). Having trained the encoder on RLScenes, we feed the representation $r^k$ of the scene as input to a control policy rewarded for solving the considered RL task.

\begin{figure*}[h]
	\centering
	\includegraphics[width=\textwidth]{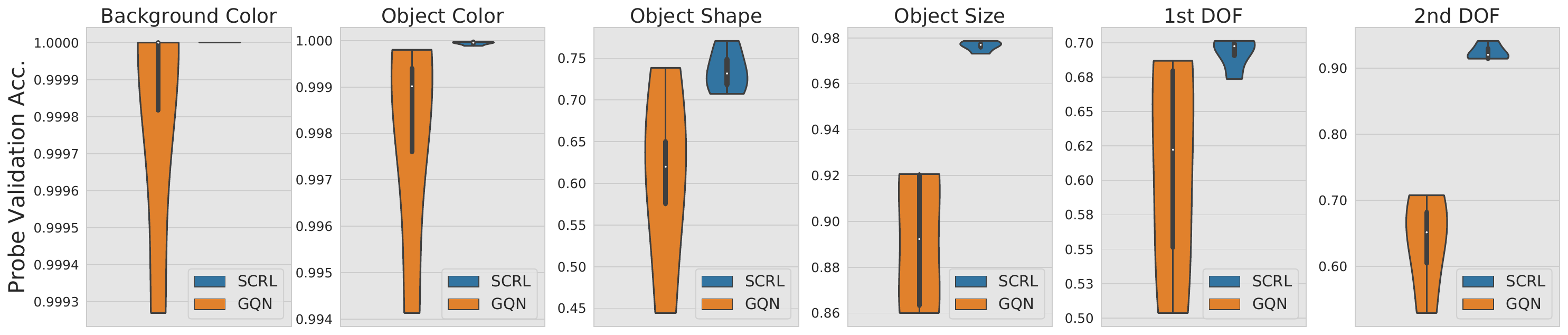}
	\caption{Quantitative Comparison of FCRL and GQN on MPI3D downstream classification tasks. The classifiers trained with FCRL's representation outperform classifiers based on GQN's representations on all the tasks.}
	\label{fig:mpi3d_downstream}
\end{figure*}%

\textbf{Scenes' Representation Learning Stage.} We use the same setting for learning the representations on both datasets. 
We fix the maximum number of context sets ($J$) to 3 in MPI3D dataset and 20 in RLScenes. The number of tuples drawn, $N$, is then chosen randomly in $[2,3]$ and $[2,20]$ respectively.

\subsubsection{Downstream Tasks on Scenes}
After training the encoder $g_{(\phi, \Phi)}$, we discard the projection head $\rho_{\phi}$ and use the trained encoder $h_{\Phi}$ to extract the representations $r^k$ and train decoders for downstream problems.%

\textbf{Scene Understanding on MPI3D Dataset.}
In MPI3D, each scene is identified by $6$ factors of variations. This allows us to define a set of $6$ tasks $\mathcal{T}_{mpi3D} = \{T^k_v\}_{v=1}^6$, where the task $T^k_v$ maps the scene to a discretized factor of variation $y^k_v$. For each task, we train a linear decoder using the objective in \Cref{eq:14}, using a single image to infer the representation $r^k$. 
\Cref{fig:mpi3d_downstream} shows the linear probes validation performance for six independently trained models on both GQN learned representations and FCRL learned representations and we see that the representations learned by FCRL consistently outperform GQN for identifying all the factors of variations in scenes.

\begin{figure}[h]
	\centering
	\includegraphics[width=0.48\textwidth]{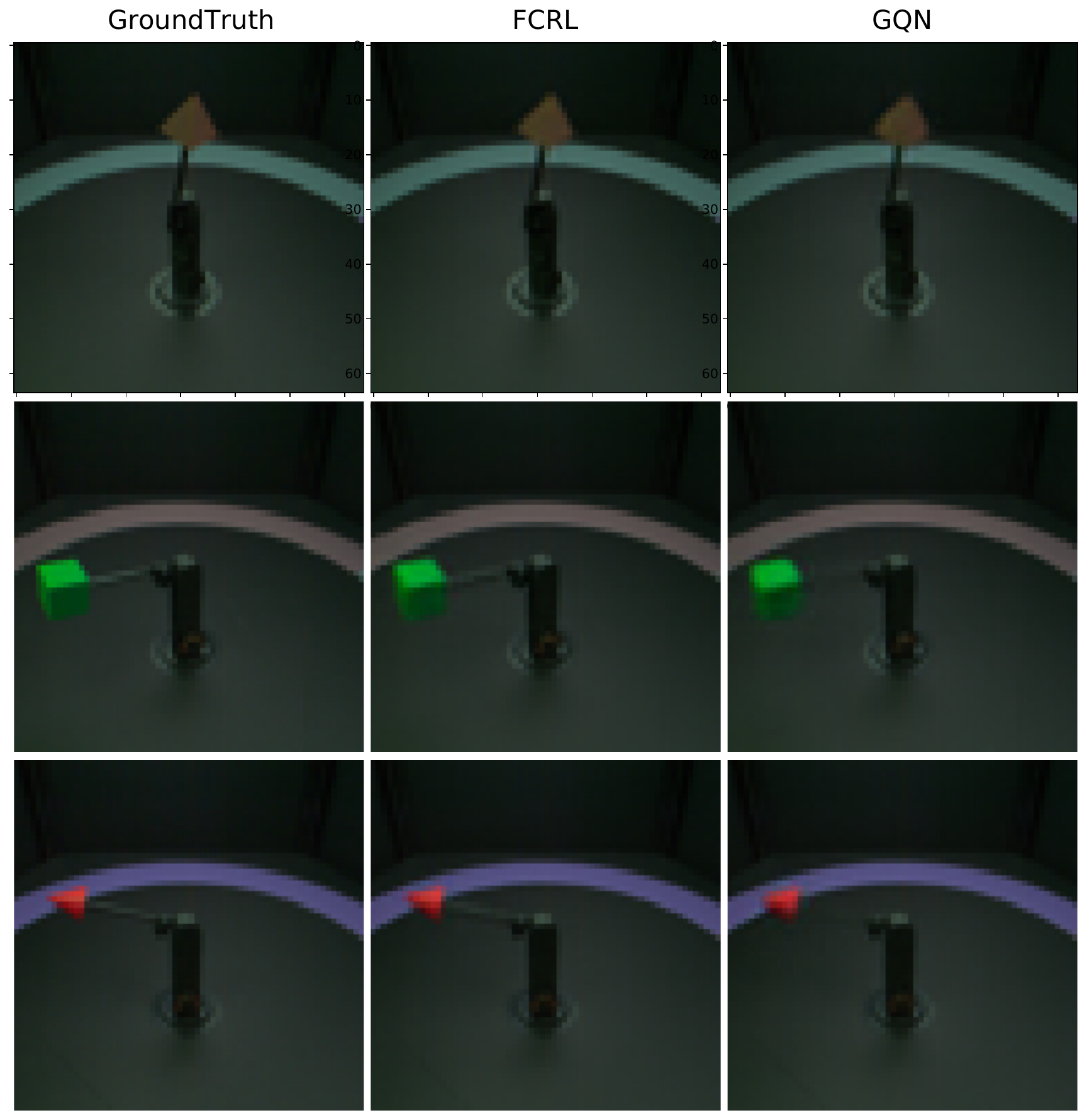}
	\caption{Qualitative Comparison of FCRL and GQN on scene reconstruction task. The decoder trained with FCRL's representation performs on par with GQN in terms of reconstructing a scene from unseen viewpoints.}
	\vspace{-2mm}
	\label{fig:mpi3d_scene_reconstruction}
\end{figure}

\textbf{Scene Reconstruction on MPI3D Dataset.} Similar to the FSIC task in \Cref{sec:2d-tasks}, we train a separate decoder to reconstruct the scenes corresponding to an unseen (query) viewpoint $x^k_q$. Conditioning on the inferred representation $r^k$ and the query viewpoint $x^k_q$, the decoder reconstructs the corresponding view of the scene $y^k_q$. The qualitative comparison in \Cref{fig:mpi3d_scene_reconstruction} shows that the decoder trained with FCRL representation is capable of preserving the information required to reconstruct the subtle details in the scene.

\begin{figure}[h]
	\centering
	\includegraphics[width=0.4\textwidth]{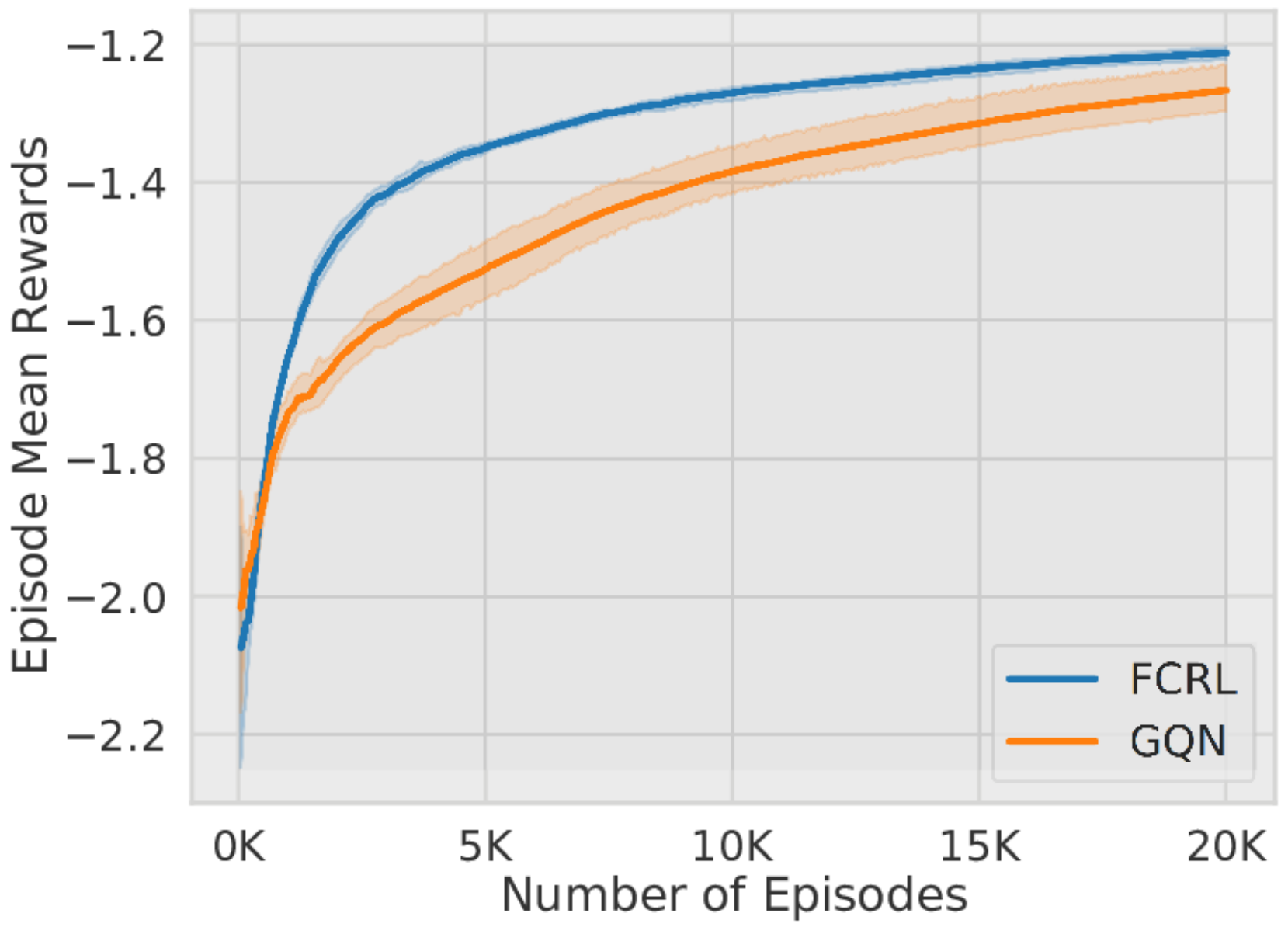}
	\caption{Comparison between GQN and FCRL on learning a data-efficient control policy for an object reaching downstream task. FCRL based policy clearly outperforms GQN based policy.} %
	\label{fig:rl}
\end{figure}

\textbf{Reinforcement Learning on RLScenes Dataset.} In RLScenes, 
the goal for the agent (a robotic finger) is to locate the object in the arena, reach it, and stay close to it for the remainder of the episode. We use the Soft-Actor-Critic (SAC) algorithm \citep{haarnoja2018soft} to learn a MLP policy for all the joints of the robot, where the policy takes as input the representation $r^k$ (inferred from a single image) and outputs an action. 
As the baseline, we use a RL policy trained with GQN representation as input. \Cref{fig:rl} shows the mean rewards and standard deviations over five runs achieved by both FCRL and GQN-based policies. We find that the FCRL agent clearly outperforms the baseline GQN-agent, both in terms of the final performance and sample-efficiency. In particular, the FCRL agent obtains convergence level control performance with approximately 2 times fewer interactions with the environment.

\section{Related Work}
\textbf{Meta-Learning.} Supervised meta-learning can be broadly classified into two main categories. The first category considers the learning algorithm to be an optimizer and meta-learning is about optimizing that optimizer, for e.g., gradient-based methods \citep{ravi2016optimization, finn2017model, li2017meta, lee2019meta} and metric-learning based methods \citep{vinyals2016matching, snell2017prototypical, sung2018learning, allen2019infinite, qiao2019transductive}. The second category is the family of Neural Processes (NP) \citep{garnelo2018conditional,garnelo2018neural,kim2019attentive, eslami2018neural} which draw inspirations from Gaussian Processes (GPs). These methods use data-specific priors in order to adapt to a new task at test time while using only a simple encoder-decoder architecture. However, they approximate the distribution over tasks in terms of their predictive distributions which does not incentivize NP to fully utilize the information in the data-specific priors. Our method draws inspiration from this simple, yet elegant framework. However, our proposed method extracts the maximum information from the context which is shown to be useful for solving not just one task, but multiple downstream tasks.

\textbf{Self-Supervised Learning.} Self-supervised learning methods aim to learn the meaningful representations of the data by performing some pretext learning tasks \citep{zhang2017split, doersch2015unsupervised}. These methods have recently received huge attention \citep{oord2018representation,tian2019contrastive,hjelm2018learning, bachman2019learning,chen2020simple,he2019momentum} mainly owing their success to noise contrastive learning objectives \citep{gutmann2010noise}. At the same time, different explanations have recently come out to explain the success of such methods for e.g. from both empirical perspective \citep{tian2020makes, tschannen2019mutual} and theoretical perspective \citep{wang2020understanding, arora2019theoretical}. The goal of these methods has mostly been to extract useful, low-dimensional representation of the data while using downstream performance as a proxy to evaluate the quality of the representation. For example, CPC \citep{oord2018representation} proposes an auto-regressive model to obtain a representation of a sample at time $t$ that is then matched with that at time $t + k$, making it specialized for sequence-valued inputs. On the other hand, \citep{tian2019contrastive,chen2020simple, he2019momentum} match the representation of a sample with the representation of its randomly augmented view. In this work, we take inspiration from these methods and propose a self-supervised learning method which meta-learn the representation of the functions. However, instead of requiring randomly augmented views or sequential ordered data points, our self-supervised loss uses partially observed, unordered views, sampled from the underlying functions.

In a similar spirit of enriching NPs \citep{garnelo2018neural} with better approximation capability, \citep{ton2019noise} proposed to replace the conditional expectations $\mathbb{E}(y|x)$ in NPs with more expressive conditional densities $p(y|x)$ estimated via NCE \citep{gutmann2010noise}. In contrast to FCRL, it directly estimates the conditional distribution $p(y|x)$ and uses a binary classifier for NCE. However, this estimation is done in a small data regime where the standard conditional density estimation does not work well. Therefore, their method is practically limited to low dimensional problems. Recently, \citep{zhang2020learning, srinivas2020curl} has shown the benefits of using self-supervised representations, learned without reconstruction, for reinforcement learning tasks. In this work, we explore the utility of such representations for the reinforcement learning tasks defined on scenes (functions).

\section{Conclusion}
In this work, we proposed a novel self-supervised representations learning algorithm for few-shot learning problems. We deviate from the commonly-used, task-specific training routines in meta-learning frameworks and propose to learn the representations of the relevant functions independently of the prediction task. Experiments on various datasets and the related set of downstream few-shot prediction tasks show the effectiveness of our method. The flexibility to reuse the same representation for different task distributions defined over functions brings us one step closer towards learning a generic meta-learning framework. Using a shared generic representation of the data-generating process, we plan to adapt the proposed framework in order to tackle multiple challenging few-shot problems such as object detection, segmentation, visual question answering.

\subsubsection*{Acknowledgments}
The authors would like to thank Krikamol Muandet, Luigi Gresele, Ilya Tolstikhin and Simon Buchholz for the helpful discussions and feedback. We thank CIFAR for their support. This work was supported by the German Federal Ministry of Education and Research (BMBF): Tübingen AI Center, FKZ: 01IS18039B, and by the Machine Learning Cluster of Excellence, EXC number 2064/1 – Project number 390727645.

\bibliography{example_paper}

\begin{thebibliography}{42}
\providecommand{\natexlab}[1]{#1}
\providecommand{\url}[1]{\texttt{#1}}
\expandafter\ifx\csname urlstyle\endcsname\relax
  \providecommand{\doi}[1]{doi: #1}\else
  \providecommand{\doi}{doi: \begingroup \urlstyle{rm}\Url}\fi

\bibitem[Allen et~al.(2019)Allen, Shelhamer, Shin, and
  Tenenbaum]{allen2019infinite}
Allen, K.~R., Shelhamer, E., Shin, H., and Tenenbaum, J.~B.
\newblock Infinite mixture prototypes for few-shot learning.
\newblock \emph{arXiv preprint arXiv:1902.04552}, 2019.

\bibitem[Anand et~al.(2019)Anand, Racah, Ozair, Bengio, C{\^o}t{\'e}, and
  Hjelm]{anand2019unsupervised}
Anand, A., Racah, E., Ozair, S., Bengio, Y., C{\^o}t{\'e}, M.-A., and Hjelm,
  R.~D.
\newblock Unsupervised state representation learning in atari.
\newblock In \emph{Advances in Neural Information Processing Systems}, pp.\
  8766--8779, 2019.

\bibitem[Arora et~al.(2019)Arora, Khandeparkar, Khodak, Plevrakis, and
  Saunshi]{arora2019theoretical}
Arora, S., Khandeparkar, H., Khodak, M., Plevrakis, O., and Saunshi, N.
\newblock A theoretical analysis of contrastive unsupervised representation
  learning.
\newblock \emph{arXiv preprint arXiv:1902.09229}, 2019.

\bibitem[Bachman et~al.(2019)Bachman, Hjelm, and
  Buchwalter]{bachman2019learning}
Bachman, P., Hjelm, R.~D., and Buchwalter, W.
\newblock Learning representations by maximizing mutual information across
  views.
\newblock In \emph{Advances in Neural Information Processing Systems}, pp.\
  15509--15519, 2019.

\bibitem[Bengio et~al.(2013)Bengio, Courville, and
  Vincent]{bengio2013representation}
Bengio, Y., Courville, A., and Vincent, P.
\newblock Representation learning: A review and new perspectives.
\newblock \emph{IEEE transactions on pattern analysis and machine
  intelligence}, 35\penalty0 (8):\penalty0 1798--1828, 2013.

\bibitem[Bousquet \& Elisseeff(2002)Bousquet and Elisseeff]{Bousquet02}
Bousquet, O. and Elisseeff, A.
\newblock Stability and generalization.
\newblock \emph{JMLR}, 2:\penalty0 499–526, 2002.

\bibitem[Chen et~al.(2020)Chen, Kornblith, Norouzi, and Hinton]{chen2020simple}
Chen, T., Kornblith, S., Norouzi, M., and Hinton, G.
\newblock A simple framework for contrastive learning of visual
  representations.
\newblock \emph{arXiv preprint arXiv:2002.05709}, 2020.

\bibitem[Doersch et~al.(2015)Doersch, Gupta, and
  Efros]{doersch2015unsupervised}
Doersch, C., Gupta, A., and Efros, A.~A.
\newblock Unsupervised visual representation learning by context prediction.
\newblock In \emph{Proceedings of the IEEE international conference on computer
  vision}, pp.\  1422--1430, 2015.

\bibitem[Doersch et~al.(2020)Doersch, Gupta, and
  Zisserman]{doersch2020crosstransformers}
Doersch, C., Gupta, A., and Zisserman, A.
\newblock Crosstransformers: spatially-aware few-shot transfer.
\newblock \emph{arXiv preprint arXiv:2007.11498}, 2020.

\bibitem[Eslami et~al.(2018)Eslami, Rezende, Besse, Viola, Morcos, Garnelo,
  Ruderman, Rusu, Danihelka, Gregor, et~al.]{eslami2018neural}
Eslami, S.~A., Rezende, D.~J., Besse, F., Viola, F., Morcos, A.~S., Garnelo,
  M., Ruderman, A., Rusu, A.~A., Danihelka, I., Gregor, K., et~al.
\newblock Neural scene representation and rendering.
\newblock \emph{Science}, 360\penalty0 (6394):\penalty0 1204--1210, 2018.

\bibitem[Finn et~al.(2017)Finn, Abbeel, and Levine]{finn2017model}
Finn, C., Abbeel, P., and Levine, S.
\newblock Model-agnostic meta-learning for fast adaptation of deep networks.
\newblock In \emph{Proceedings of the 34th International Conference on Machine
  Learning-Volume 70}, pp.\  1126--1135. JMLR. org, 2017.

\bibitem[Garnelo et~al.(2018{\natexlab{a}})Garnelo, Rosenbaum, Maddison,
  Ramalho, Saxton, Shanahan, Teh, Rezende, and Eslami]{garnelo2018conditional}
Garnelo, M., Rosenbaum, D., Maddison, C.~J., Ramalho, T., Saxton, D., Shanahan,
  M., Teh, Y.~W., Rezende, D.~J., and Eslami, S.
\newblock Conditional neural processes.
\newblock \emph{arXiv preprint arXiv:1807.01613}, 2018{\natexlab{a}}.

\bibitem[Garnelo et~al.(2018{\natexlab{b}})Garnelo, Schwarz, Rosenbaum, Viola,
  Rezende, Eslami, and Teh]{garnelo2018neural}
Garnelo, M., Schwarz, J., Rosenbaum, D., Viola, F., Rezende, D.~J., Eslami, S.,
  and Teh, Y.~W.
\newblock Neural processes.
\newblock \emph{arXiv preprint arXiv:1807.01622}, 2018{\natexlab{b}}.

\bibitem[Gondal et~al.(2019)Gondal, Wuthrich, Miladinovic, Locatello, Breidt,
  Volchkov, Akpo, Bachem, Sch{\"o}lkopf, and Bauer]{gondal2019transfer}
Gondal, M.~W., Wuthrich, M., Miladinovic, D., Locatello, F., Breidt, M.,
  Volchkov, V., Akpo, J., Bachem, O., Sch{\"o}lkopf, B., and Bauer, S.
\newblock On the transfer of inductive bias from simulation to the real world:
  a new disentanglement dataset.
\newblock In \emph{Advances in Neural Information Processing Systems}, pp.\
  15740--15751, 2019.

\bibitem[Gordon et~al.(2020)Gordon, Bruinsma, Foong, Requeima, Dubois, and
  Turner]{gordon2020convolutional}
Gordon, J., Bruinsma, W., Foong, A.~Y., Requeima, J., Dubois, Y., and Turner,
  R.~E.
\newblock Convolutional conditional neural processes.
\newblock 2020.

\bibitem[Gutmann \& Hyv{\"a}rinen(2010)Gutmann and
  Hyv{\"a}rinen]{gutmann2010noise}
Gutmann, M. and Hyv{\"a}rinen, A.
\newblock Noise-contrastive estimation: A new estimation principle for
  unnormalized statistical models.
\newblock In \emph{Proceedings of the Thirteenth International Conference on
  Artificial Intelligence and Statistics}, pp.\  297--304, 2010.

\bibitem[Haarnoja et~al.(2018)Haarnoja, Zhou, Abbeel, and
  Levine]{haarnoja2018soft}
Haarnoja, T., Zhou, A., Abbeel, P., and Levine, S.
\newblock Soft actor-critic: Off-policy maximum entropy deep reinforcement
  learning with a stochastic actor.
\newblock \emph{arXiv preprint arXiv:1801.01290}, 2018.

\bibitem[He et~al.(2019)He, Fan, Wu, Xie, and Girshick]{he2019momentum}
He, K., Fan, H., Wu, Y., Xie, S., and Girshick, R.
\newblock Momentum contrast for unsupervised visual representation learning.
\newblock \emph{arXiv preprint arXiv:1911.05722}, 2019.

\bibitem[Hjelm et~al.(2018)Hjelm, Fedorov, Lavoie-Marchildon, Grewal, Bachman,
  Trischler, and Bengio]{hjelm2018learning}
Hjelm, R.~D., Fedorov, A., Lavoie-Marchildon, S., Grewal, K., Bachman, P.,
  Trischler, A., and Bengio, Y.
\newblock Learning deep representations by mutual information estimation and
  maximization.
\newblock \emph{arXiv preprint arXiv:1808.06670}, 2018.

\bibitem[Joshi et~al.(2020)Joshi, Widmaier, Agrawal, and
  Wüthrich]{trifinger-simulation}
Joshi, S., Widmaier, F., Agrawal, V., and Wüthrich, M.
\newblock
  \url{https://github.com/open-dynamic-robot-initiative/trifinger_simulation},
  2020.

\bibitem[Kim et~al.(2019)Kim, Mnih, Schwarz, Garnelo, Eslami, Rosenbaum,
  Vinyals, and Teh]{kim2019attentive}
Kim, H., Mnih, A., Schwarz, J., Garnelo, M., Eslami, A., Rosenbaum, D.,
  Vinyals, O., and Teh, Y.~W.
\newblock Attentive neural processes.
\newblock \emph{arXiv preprint arXiv:1901.05761}, 2019.

\bibitem[Kipf et~al.(2019)Kipf, van~der Pol, and Welling]{kipf2019contrastive}
Kipf, T., van~der Pol, E., and Welling, M.
\newblock Contrastive learning of structured world models.
\newblock \emph{arXiv preprint arXiv:1911.12247}, 2019.

\bibitem[LeCun et~al.(1998)LeCun, Bottou, Bengio, and
  Haffner]{lecun1998gradient}
LeCun, Y., Bottou, L., Bengio, Y., and Haffner, P.
\newblock Gradient-based learning applied to document recognition.
\newblock \emph{Proceedings of the IEEE}, 86\penalty0 (11):\penalty0
  2278--2324, 1998.

\bibitem[Lee et~al.(2019)Lee, Maji, Ravichandran, and Soatto]{lee2019meta}
Lee, K., Maji, S., Ravichandran, A., and Soatto, S.
\newblock Meta-learning with differentiable convex optimization.
\newblock In \emph{Proceedings of the IEEE Conference on Computer Vision and
  Pattern Recognition}, pp.\  10657--10665, 2019.

\bibitem[Li et~al.(2017)Li, Zhou, Chen, and Li]{li2017meta}
Li, Z., Zhou, F., Chen, F., and Li, H.
\newblock Meta-sgd: Learning to learn quickly for few-shot learning.
\newblock \emph{arXiv preprint arXiv:1707.09835}, 2017.

\bibitem[Oord et~al.(2018)Oord, Li, and Vinyals]{oord2018representation}
Oord, A. v.~d., Li, Y., and Vinyals, O.
\newblock Representation learning with contrastive predictive coding.
\newblock \emph{arXiv preprint arXiv:1807.03748}, 2018.

\bibitem[Qiao et~al.(2019)Qiao, Shi, Li, Wang, Huang, and
  Tian]{qiao2019transductive}
Qiao, L., Shi, Y., Li, J., Wang, Y., Huang, T., and Tian, Y.
\newblock Transductive episodic-wise adaptive metric for few-shot learning.
\newblock In \emph{Proceedings of the IEEE International Conference on Computer
  Vision}, pp.\  3603--3612, 2019.

\bibitem[Ravi \& Larochelle(2016)Ravi and Larochelle]{ravi2016optimization}
Ravi, S. and Larochelle, H.
\newblock Optimization as a model for few-shot learning.
\newblock 2016.

\bibitem[Snell et~al.(2017)Snell, Swersky, and Zemel]{snell2017prototypical}
Snell, J., Swersky, K., and Zemel, R.
\newblock Prototypical networks for few-shot learning.
\newblock In \emph{Advances in neural information processing systems}, pp.\
  4077--4087, 2017.

\bibitem[Srinivas et~al.(2020)Srinivas, Laskin, and Abbeel]{srinivas2020curl}
Srinivas, A., Laskin, M., and Abbeel, P.
\newblock Curl: Contrastive unsupervised representations for reinforcement
  learning.
\newblock \emph{arXiv preprint arXiv:2004.04136}, 2020.

\bibitem[Sung et~al.(2018)Sung, Yang, Zhang, Xiang, Torr, and
  Hospedales]{sung2018learning}
Sung, F., Yang, Y., Zhang, L., Xiang, T., Torr, P.~H., and Hospedales, T.~M.
\newblock Learning to compare: Relation network for few-shot learning.
\newblock In \emph{Proceedings of the IEEE Conference on Computer Vision and
  Pattern Recognition}, pp.\  1199--1208, 2018.

\bibitem[Tian et~al.(2019)Tian, Krishnan, and Isola]{tian2019contrastive}
Tian, Y., Krishnan, D., and Isola, P.
\newblock Contrastive multiview coding.
\newblock \emph{arXiv preprint arXiv:1906.05849}, 2019.

\bibitem[Tian et~al.(2020)Tian, Sun, Poole, Krishnan, Schmid, and
  Isola]{tian2020makes}
Tian, Y., Sun, C., Poole, B., Krishnan, D., Schmid, C., and Isola, P.
\newblock What makes for good views for contrastive learning.
\newblock \emph{arXiv preprint arXiv:2005.10243}, 2020.

\bibitem[Ton et~al.(2019)Ton, Chan, Teh, and Sejdinovic]{ton2019noise}
Ton, J.-F., Chan, L., Teh, Y.~W., and Sejdinovic, D.
\newblock Noise contrastive meta-learning for conditional density estimation
  using kernel mean embeddings.
\newblock \emph{arXiv preprint arXiv:1906.02236}, 2019.

\bibitem[Tschannen et~al.(2019)Tschannen, Djolonga, Rubenstein, Gelly, and
  Lucic]{tschannen2019mutual}
Tschannen, M., Djolonga, J., Rubenstein, P.~K., Gelly, S., and Lucic, M.
\newblock On mutual information maximization for representation learning.
\newblock \emph{arXiv preprint arXiv:1907.13625}, 2019.

\bibitem[Vapnik(1995)]{Vapnik95}
Vapnik, V.
\newblock \emph{The Nature of Statistical Learning Theory}.
\newblock Springer, NY, 1995.

\bibitem[Vinyals et~al.(2016)Vinyals, Blundell, Lillicrap, Wierstra,
  et~al.]{vinyals2016matching}
Vinyals, O., Blundell, C., Lillicrap, T., Wierstra, D., et~al.
\newblock Matching networks for one shot learning.
\newblock In \emph{Advances in neural information processing systems}, pp.\
  3630--3638, 2016.

\bibitem[Wang \& Isola(2020)Wang and Isola]{wang2020understanding}
Wang, T. and Isola, P.
\newblock Understanding contrastive representation learning through alignment
  and uniformity on the hypersphere.
\newblock \emph{arXiv preprint arXiv:2005.10242}, 2020.

\bibitem[Xu et~al.(2019)Xu, Ton, Kim, Kosiorek, and Teh]{xu2019metafun}
Xu, J., Ton, J.-F., Kim, H., Kosiorek, A.~R., and Teh, Y.~W.
\newblock Metafun: Meta-learning with iterative functional updates.
\newblock \emph{arXiv preprint arXiv:1912.02738}, 2019.

\bibitem[Zaheer et~al.(2017)Zaheer, Kottur, Ravanbakhsh, Poczos, Salakhutdinov,
  and Smola]{zaheer2017deep}
Zaheer, M., Kottur, S., Ravanbakhsh, S., Poczos, B., Salakhutdinov, R.~R., and
  Smola, A.~J.
\newblock Deep sets.
\newblock In \emph{Advances in neural information processing systems}, pp.\
  3391--3401, 2017.

\bibitem[Zhang et~al.(2020)Zhang, McAllister, Calandra, Gal, and
  Levine]{zhang2020learning}
Zhang, A., McAllister, R., Calandra, R., Gal, Y., and Levine, S.
\newblock Learning invariant representations for reinforcement learning without
  reconstruction.
\newblock \emph{arXiv preprint arXiv:2006.10742}, 2020.

\bibitem[Zhang et~al.(2017)Zhang, Isola, and Efros]{zhang2017split}
Zhang, R., Isola, P., and Efros, A.~A.
\newblock Split-brain autoencoders: Unsupervised learning by cross-channel
  prediction.
\newblock In \emph{Proceedings of the IEEE Conference on Computer Vision and
  Pattern Recognition}, pp.\  1058--1067, 2017.

\end{thebibliography}
\bibliographystyle{icml2021}

\newpage

\appendix
\section{Appendix}

\subsection{Ablation Study}
\label{sec:ablations_appendix}

\begin{figure*}%
	\centering
	\includegraphics[width=\textwidth]{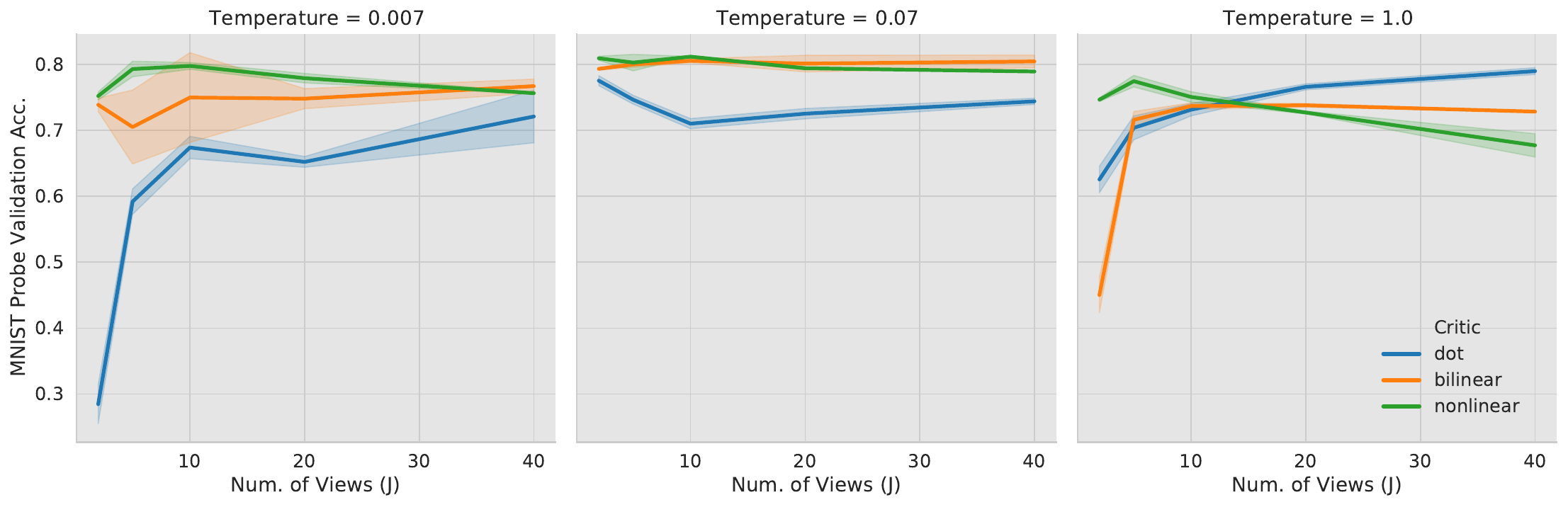}
	\caption{An ablation study for understanding the role of the number of partial observations ($J$), the critics and the temperature parameter ($\tau$) for learning FCRL based representation. The graph shows the accuracy achieved by the linear classifier (Few-shot content classification task) on MNIST digits validation dataset. Note that each image corresponds to a 2D functions, and the accuracy bands show the standard deviation of three independent runs.}
	\label{fig:mnist_fcrl_ablation}
\end{figure*}%
\paragraph{Role of Number of Observations (J)}
The number of observations $J$ corresponds to the number of partial observations that we have of a functions $f^k$. Ideally, we only need two such observations to learn the representations via contrastive objective. However, it has been shown that having more positive pairs result in learning better representations \citep{chen2020simple, tian2019contrastive}.

It should be noted that in our setting, the number of observed context sets $N$ do not necessarily correspond to the number of observations $J$. This is because for some datasets, we aggregate the context points to get one partial observation (see \Cref{fig:loss}). This is important for the simple 1D and 2D regression functions where a single context point does not provide much information, hence the individual partial observations need more than one context point. In our setting, we treat $J$ as a hyperparameter whose optimal value varies depending on the function. For instance, the MPI3D scene dataset has only $3$ views per scene, therefore $J$ can not be greater than $3$ and we keep it fixed. For 1D and 2D regression functions, we observe that for a fixed number of context points $N$, the optimal number of observations $J$ varies. For understanding the role of $J$ better in these experiments, we perform an ablation study on MNIST2D dataset with FSCC (few-shot content classification) as the downstream task \Cref{fig:mnist_fcrl_ablation}. For each hyper-parameter configuration we train three models, initialized with different random seeds. The maximum number of context points is fixed to $200$ while the $J$ varies between $2$ and $40$.

It can be seen that the smaller values of $J$ do not result in better FSCC score, however, the accuracy seems to plateau after $J=10$. Therefore, we fix it to $10$ for MNIST2D. For RLScene dataset, we fixed the maximum number of context points to be $20$ and found the optimal number of partial observations to be $J=4$. Note that the FSCC accuracy seems to be more influenced by the hyperparameters of critic and temperature, shown concurrently in the \Cref{fig:mnist_fcrl_ablation}. We discuss their roles in the next section.

\paragraph{Role of Critics and Temperature $\tau$.}
We regard the discriminative scoring functions, including the projection heads, as critics. The simplest critic function does not contain any projection layer, regarded as \emph{dot product} critic, where the contrastive objective is defined directly on the representations returned by the base encoder $h_{\Phi}$. However, recently the role of critics in learning better representations has been explored in depth \citep{oord2018representation, chen2020simple}. Building on these findings, we evaluate the role played by different critics in learning the functions representations. \Cref{fig:mnist_fcrl_ablation} shows the ablation for three different critics on MNIST2D validation dataset. It can be seen that the performance of critics is also highly linked with the temperature parameter $\tau$. For an optimal temperature value $\tau$, the non-linear critic performs consistently better.

\begin{figure*}[h]
	\centering
	\includegraphics[width=\textwidth]{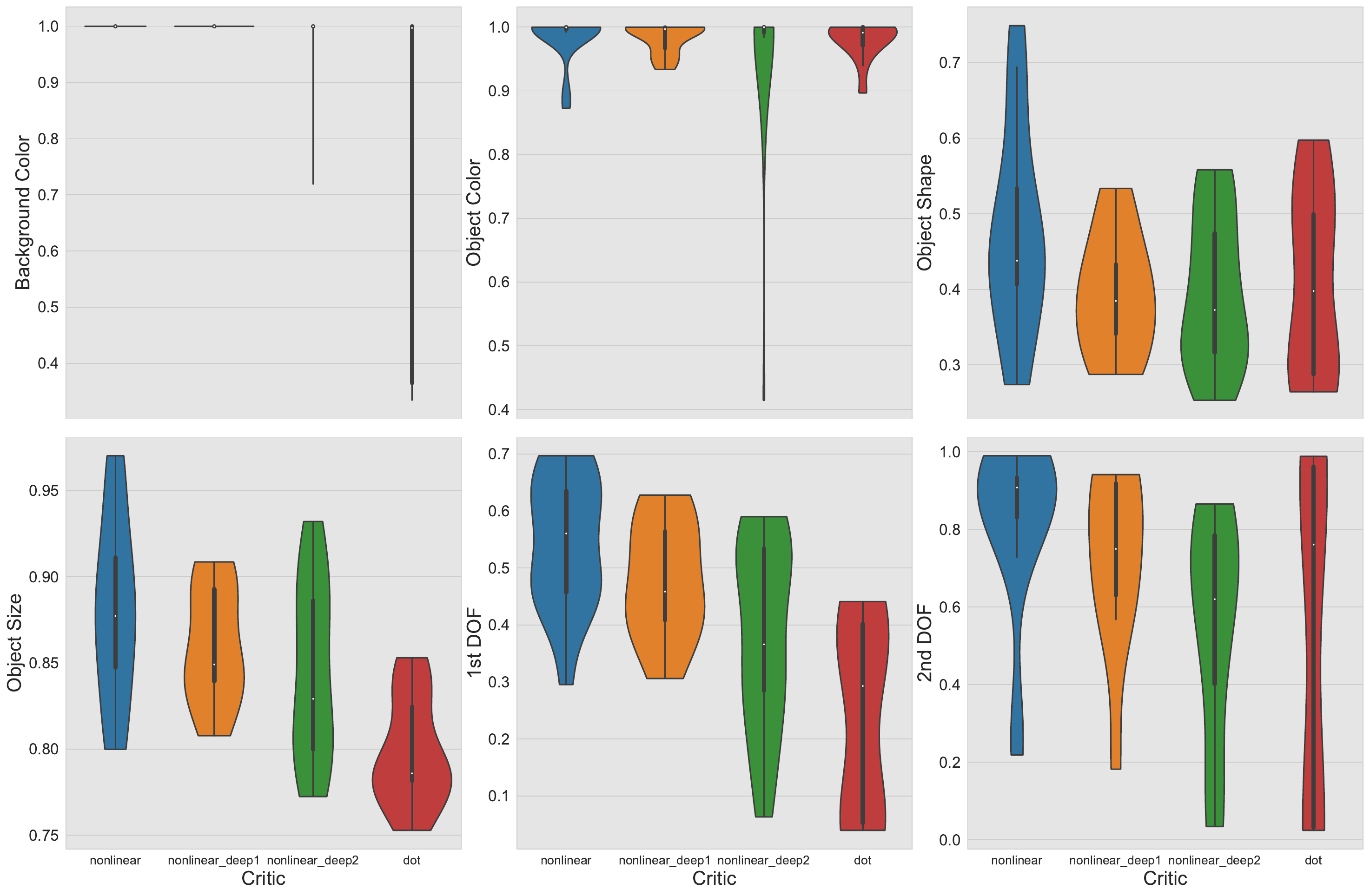}
	\caption{An ablation study to understand the role of different critics in learning meta-representations of MPI3D scenes via FCRL. Shown here is the accuracy achieved by six different linear classifiers, trained on the representations, for identifying the six factors of variation. Non-linear critic consistently performs better than the other critics.}
	\label{fig:mpi3d_ablation}
\end{figure*}

Such hyperparameter grid search (done for MNIST2D) is very expensive for the ablation studies on the bigger datasets, such as, the MPI3D and RLScenes datasets. Therefore, we define the range for the $t$ and perform a random sweep of 80 models with randomly selected hyper-parameter values for critic and temperature on MPI3D dataset. We did not find any pattern for the effect of temperature $\tau$ on the downstream tasks, however the pattern emerged for the class of critics. \Cref{fig:mpi3d_ablation} shows the ablation for critics on MPI3D dataset. It can be seen that the non-linear critic performs better in extracting features which are useful for the downstream classification tasks. Because of this trend across two different datasets, we therefore performed all our experiments with non-linear critic. The project head in nonlinear critics is defined as an MLP with one hidden layer and batch normalization in between.
\begin{figure}[h]
	\centering
	\includegraphics[width=0.46\textwidth]{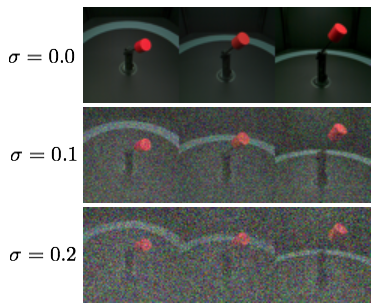}
	\caption{MPI3D dataset with the varying level of additive Gaussian noise.}
	\label{fig:mpi3d_noisy_appendix}
\end{figure}%

\section{Robustness to Noise}
\label{sec:noise_robustness_appendix}
\begin{figure*}%
	\centering
	\includegraphics[width=\textwidth]{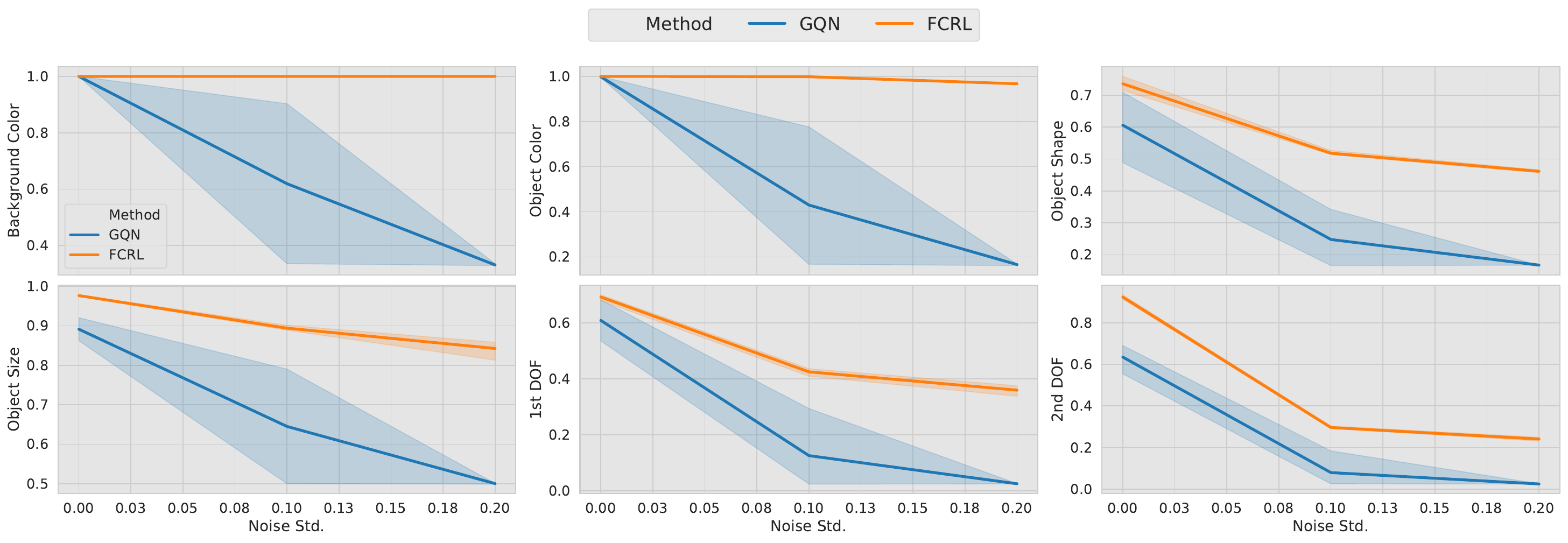}
	\caption{Quantitative comparison of FCRL and GQN for noise robustness on MPI3D downstream tasks. X-axis in each plot corresponds to the varying level of Gaussian noise (as depicted in \Cref{fig:mpi3d_noisy_appendix}). The representations are extracted from one view only, and the accuracy bands show the standard deviation of five independent runs. It can be seen that the linear classifiers trained on FCRL representations are able to classify the digits even in extremely noisy case, significantly outperforming the classifiers trained on GQN representation.
}
	\label{fig:mpi3d_downstream_noise}
\end{figure*}

Functions with additive noise have been well-studied, however, the contemporary literature in meta-learning mostly considers them to be noise free. In this work, we explore whether the learned representations of the functions are prone to the additive Gaussian noise. We consider the form of the function as given in \Cref{eq:func} and vary the standard deviation of the added noise. It can be seen that with the increased level of noise the features in the image start to diminish, shown in the \Cref{fig:mpi3d_noisy_appendix}. GQN approach the learning problem by reconstructing these noisy images where the signal to noise ratio is very low. On the other hand, FCRL learns to contrast the scene representations with other scenes without requiring any reconstruction in the pixel space. This helps it in extracting invariant features in the views of a scene, getting rid of any non-correlated noise in the input.\\
In addition to the analysis on MNIST 2D regression task in \Cref{figure:mnist_probe_noise}, we test the performance of these representations learning algorithm on MPI3D factors identification task in \Cref{fig:mpi3d_downstream_noise}. It can be seen that FCRL representations can recover the information about the data factors, even in the extreme case where the noise level is very high (standard deviation of 0.2). Whereas, GQN performs very poorly such that the downstream probes achieve the random accuracy.

\section{Estimating Density Ratios corresponding to the Functions}
\label{sec:optimal}

The contrastive objective in \Cref{eq:2}, in essence, tries to solve a classification problem i.e. to identify whether the given observation $O^i$ comes from the function $f^i$ or not. The supervision signal is provided by taking another observation $\hat{O}$ from the same function $f^i$ as an anchor (a target label), thus making it a self-supervised method. This self-supervised, view-classification task, for a function $f^i$, leads to the estimation of density ratios between the joint distribution of observations $p(O^1, O^2 | i)$ and their product of marginals $p(O^1|i)p(O^2|i)$. This joint distribution in turn corresponds to the joint distribution of the input-output pairs of the function $p(x,y|i)$. This way of learning a function's distribution is different from the typical regression objectives, which learn about a given function $f^i$ by trying to approximate the predictive distribution $p(y|x)$.

By assuming the universal function approximation capability of $g_{(\phi, \Phi)}$, and the availability of infinitely many functions $f^k \sim p(f)$ with fixed number of context points $N$ each, the model posterior learned by the optimal classifier corresponding to \Cref{eq:2} would be equal to the true posterior given by Bayes rule. 
\begin{align}
f^{k} & \sim P(f)\quad\forall \; \; k\in\{1,..,K\}\\
O^{k} & \sim P(O|f^{k})\quad\forall k\in\{1,..,K\}\\
i & \sim\mathcal{U}(K)\\
\hat{f} & =f^{i}\\
\hat{O} & \sim P(O|\hat{f})\\
p(i|O^{1:K},\hat{O}) & =\frac{p(O^{1:K},\hat{O}|i)p(i)}{\sum_{i}p(O^{1:K},\hat{O}|i)p(i)}\\
 & =\frac{p(O^{i},\hat{O}|i)p(i)\prod_{k\neq i}p(O^{k}|i)p(\hat{O}|i)}{\sum_{j}p(O^{j},\hat{O}|j)p(j)\prod_{k\neq j}p(O^{k}|j)p(\hat{O}|j)}\\
& =\frac{\frac{p(O^{i},\hat{O}|i)}{p(O_{i})p(\hat{O})}p(i)}{\sum_{j}\frac{p(O^{j},\hat{O}|j)}{p(O_{j})p(\hat{O})}p(j)}
\end{align}

The posterior probability for a function $f^i$ is proportional to the class-conditional probability density function $p(O^{i},\hat{O} | i)$, which shows the probability of observing the pair $(O^i,\hat{O})$ from function $f^i$. The optimal classifier would then be proportional to the density ratio given below
\begin{equation}
\exp (sim_{(\phi, \Phi)}(\hat{O}, O^i)) \propto  \frac{p(O^{i},\hat{O})}{p(O_{i})p(\hat{O})}
\end{equation}
Similar analysis has been shown by the \citep{oord2018representation} for showing the mutual information perspective associated with self-supervised contrastive objective (infoNCE). The joint distribution over the pair of observations correspond to the distribution of the underlying function $f^i$. Thus, given some observation of a function, an optimal classifier would attempt at estimating the true density of the underlying function.

\section{Scenes' Datasets}
\label{sec:scenes_datasets_appendix}
\begin{figure*}[h]
\begin{subfigure}{0.37\textwidth}
\centering\includegraphics[width=\textwidth]{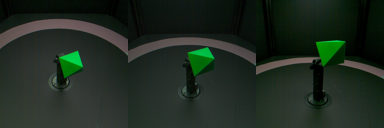}
\caption{}
\end{subfigure} %
\begin{subfigure}{0.62\textwidth}
\centering\includegraphics[width=\textwidth]{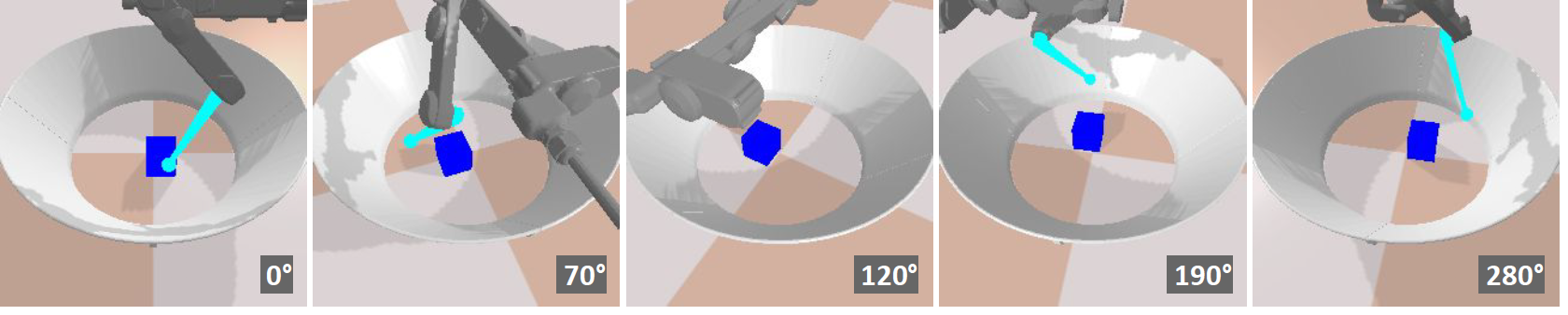}
\caption{}
\end{subfigure} \hspace{-0.7cm} %
\caption{Datasets for Scenes Representation Learning (a) MPI3D \citep{gondal2019transfer} has three camera viewpoints, with images of a robotics arm manipulating an object. (b) RLScenes has $36$ possible camera viewpoints for capturing an arena consisting of a robot finger and an object.}
\label{fig:scene_datasets}
\vspace{-3mm}
\end{figure*}

\paragraph{MPI3D Dataset.} The MPI3D dataset \citep{gondal2019transfer} is introduced to study transfer in unsupervised representations learning algorithms. The dataset comes in three different formats, varying in the levels of realism i.e. real-world, simulated-realistic and simulated-toy. Each dataset contains
$1,036,800$ images of a robotic manipulator each, encompassing seven different factors of variations i.e., object colors ($6$ values), object shapes ($6$ values), object sizes ($2$ values), camera heights ($3$ values), background colors ($3$ values), rotation along first degree of freedom (($40$ values)) and second degree of freedom (($40$ values)). Thus, each image represents a unique combination of all the factors. See \Cref{fig:scene_datasets}(a) to see a sample of the dataset.

\noindent In this work, we consider the real-world version of the dataset. The multi-view setting is formulated by considering the images of a scene captured by three different cameras, placed at different heights. This effectively gives us $345,600$ scenes with three views each. We split the dataset into training and validation chunks, where the training dataset contains $310,000$ scenes and the validation dataset contains the rest $35,600$ scenes, approximately $10\%$ of the dataset.

\paragraph{RLScenes.} The RLScenes dataset is generated in simulation using \citep{trifinger-simulation} for a single 3-DOF robotic manipulator in a 3D environment. The dataset consists of $40,288$ scenes, each scene parametrized by: object colors (one of $4$), robot tip colours (one of $3$), robot positions (uniformly sampled from the range of feasible joint values), and object positions (uniformly sampled within an arena bounded by a high boundary as seen in \Cref{fig:scene_datasets}). Each scene consists of $36$ views, corresponding to the uniformly distributed camera viewpoints along a ring of fixed radius and fixed height, defined above the environment. As can be seen in \Cref{fig:scene_datasets}, the robot finger might not be visible completely in all the views, or the object might be occluded in some view. The $36$ views help by capturing a $360 \deg$ holistic perspective of the environment. First a configuration of the above scene parameters is selected and displayed in the scene, then the camera is revolved along the ring to capture its multiple views. For learning the scene representations via both FCRL and GQN, we split the dataset into $35000$ training and $5288$ validation points. 

\section{Details of Experiments on Scene Representation Learning}
\label{sec:scenes_experiments_appendix}
\begin{figure*}[h]
	\centering
	\includegraphics[width=\textwidth]{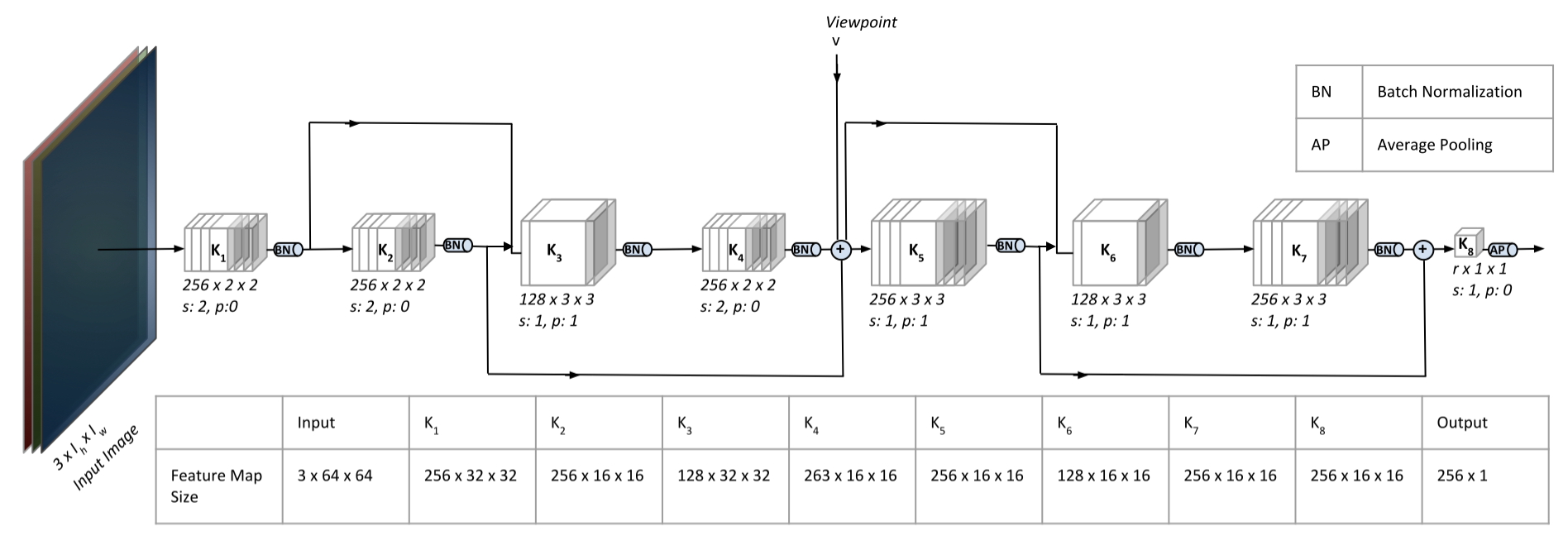}
	\caption{The "pool" architecture used for learning representations for the scenes' datasets via FCRL. The architecture is the same as was used for training GQN \citep{eslami2018neural}.}
	\label{fig:net_arch}
\end{figure*}

For learning the scene representations for both MPI3D dataset and RL Scenes, we used similar base encoder architecture. More specifically, we adapted the \lq\lq pool" architecture provided in GQN \citep{eslami2018neural}, as it has been regarded to exhibit better view-invariance, factorization and compositional characteristics as per the comprehensive study done in \citep{eslami2018neural}. We further augmented this architecture with batch-normalization. The architecture we use is shown in \Cref{fig:net_arch}:\\

\subsection{Learning Scenes for MPI3D Dataset.}
Since we have three view per each scene in MPI3D dataset, therefore we are restricted in defining the number of context points and the number of views. In all our experiments the number of context points $N$ is fixed to three while the number of views $J$ is set to two. In contrast to the experiments for regression datasets where more datapoints per each view resulted in better representations, the restriction to a single image view in MPI3D dataset did not hurt the representation quality, measured in terms of downstream performance. In the downstream experiments on MPI3D, we use only one image to train and validate the probes. Due to the limitation of available views, we could not measure the effect of varying the number of views on the downstream performance.\\

Even though no explicit structural constraint was imposed for learning the representation. The FCRL algorithm implicitly figures out the commonality between the factors in different scenes. We visualize these latent clusterings in the \Cref{fig:mpi3d_latents}. We plot the 2D TSNE embeddings of the 128D representations inferred by the model. Thereafter, to visualize the structure corresponding to each factor, we only vary one factor and fix the rest of them except for first degree of freedom and second degree of freedom factors. A clear structure can be seen in the learned representations.

\paragraph{Implementation Details.}
We use the GQNs \lq pool' architecture with batch normalization as encoder. As mentioned in the ablation study, we did a random sweep over the range of hyperparameters and selected the best performing model. Further details on the hyperparameters is provided in \Cref{table:encoder_scene_params}.

\subsection{Learning Scenes for RLScenes Dataset.}
To train the FCRL encoder, we randomly sample the number of views from each scene to lie within the range [2, 20]: upper-bounded by 20 to restrict the maximum number of images per scene to be the same as that used in \citep{eslami2018neural}, and lower bounded by 2 in case just the one view is not from a suitable angle. So, here, the maximum number of context points $N$ is $20$. The number of subsets $J$ is set to $8$. In the downstream reinforcement learning task, we use only one image to train the policy network, as is the usual practice, and the same as \citep{eslami2018neural}. We kept the joint ranges from which joint positions are uniformly sampled to randomly reset the robot at the beginning of every episode while training the policy network to be the same as the ranges used for sampling the robot position while generating the dataset to train the FCRL encoder. These joint ranges are selected so as to ensure that there are more scenes in which the robot finger is visible. However, in order to not constrain the agent's exploration, we let the action space for training the reaching agent to be less constrained, and be able to explore the entire range from $-pi$ to $pi$. So, effectively, the space seen by the robot during the training of the representations is a subspace of that seen while inferring the representations from the environment used for this downstream reaching task. Interestingly, the inferred representations can also work effectively on unseen robot configurations as demonstrated by the success of the reacher. 

\paragraph{Implementation Details.}
Similar to the encoder training for MPI3D scenes, we learned the encoder for RLScenes. However, since the downstream tasks is a reinforcement learning task, it was hard to judge the quality of representations. Therefore, we took some insights from the MPI3D experiments and selected the model, trained with hyperparameters, which performed the best on the RL downstream tasks. Further details on the hyperparameters is provided in \Cref{table:encoder_scene_params}.

\begin{table*}[]\caption{Hyperparameters Settings for Scene Representation Learning Experiments.}\label{table:encoder_scene_params}
\centering
\begin{subtable}[t]{0.5\linewidth}
\vspace{2mm}
    \begin{tabular}{l  l}
      \toprule
      \textbf{Parameter} & \textbf{Values}\\
      \midrule 
      Batch size & $64$\\
      Representation dimension & $128$\\
      Temperature: $\tau$ & 0.88 \\
      Number of subsets: $J$ & 2 \\
      Max number of context points: 3\\
      Epochs & 100 \\
      Critic & Nonlinear \\
      Objective & NCE \\
      Optimizer & Adam\\
      Adam: beta1 & 0.9\\
      Adam: beta2 & 0.999\\
      Adam: epsilon & 1e-8\\
      Adam: learning rate & 0.0005\\
      Learning Rate Scheduler & Cosine \\ 
      Number of workers & 10 \\
      Batch normalization & Yes \\
      \bottomrule
    \end{tabular}
    \vspace{0.5cm}
    \caption{Hyperparameters to train FCRL based encoder on the MPI3D Dataset.}
\end{subtable}%
\hspace{5mm}
\begin{subtable}[t]{0.4\linewidth}
\vspace{2mm}
    \begin{tabular}{l  l}
      \toprule
      \textbf{Parameter} & \textbf{Values}\\
      \midrule 
      Batch size & $32$\\
      Representation dimension & $256$\\
      Temperature: $\tau$ & 0.46 \\
      Number of subsets: $J$ & 4 \\
      Max number of context points: 20\\
      Epochs & 100 \\
      Critic & Nonlinear \\
      Objective & NCE \\
      Optimizer & Adam\\
      Adam: beta1 & 0.9\\
      Adam: beta2 & 0.999\\
      Adam: epsilon & 1e-8\\
      Adam: learning rate & 0.0005\\
      Learning Rate Scheduler & Cosine \\ 
      Number of workers & 10 \\
      Batch normalization & Yes \\
      \bottomrule
    \end{tabular}
    \vspace{0.5cm}
    \caption{Hyperparameters to train FCRL based encoder on the RLScenes Dataset.}
\end{subtable}%
\end{table*}

\section{Details of Experiments on 1D Functions}
\label{sec:exp-1D}

\begin{table*}[]\caption{Hyperparameters Settings for Sinusoid Experiments.}\label{table:sinusoid_params}
\centering
\begin{subtable}[t]{0.5\linewidth}
\vspace{2mm}
    \begin{tabular}{l  l}
      \toprule
      \textbf{Parameter} & \textbf{Values}\\
      \midrule 
      Batch size & $256$\\
      Latent space dimension & $50$\\
      Temperature: $\tau$ & 0.07 \\
      Number of subsets: $J$ & 2 \\
      Max number of context points: $N$ & 20\\
      Epochs & 30 \\
      Critic & Nonlinear \\
      Optimizer & Adam\\
      Adam: beta1 & 0.9\\
      Adam: beta2 & 0.999\\
      Adam: epsilon & 1e-8\\
      Adam: learning rate & 0.0003\\
      Learning Rate Scheduler & Cosine \\ 
      \bottomrule
    \end{tabular}
    \vspace{0.5cm}
    \caption{Hyperparameters to train FCRL based encoder for 1D sinusoid functions.}
\end{subtable}%
\hspace{5mm}
\begin{subtable}[t]{0.4\linewidth}
\vspace{2mm}
    \begin{tabular}{l  l}
      \toprule
      \textbf{Parameter} & \textbf{Values}\\
      \midrule 
      Batch size & $256$\\
      Epochs & 30 \\
      Critic & Nonlinear \\
      Optimizer & Adam\\
      Adam: beta1 & 0.9\\
      Adam: beta2 & 0.999\\
      Adam: epsilon & 1e-8\\
      Adam: learning rate & 0.001\\
      Learning Rate Scheduler & Cosine \\ 
      \bottomrule
    \end{tabular}
    \vspace{0.5cm}
    \caption{Hyperparameters to train FSR Decoder on FCRL learned representations.}
\end{subtable}%
\end{table*}

\paragraph{Implementation Details.}
We used the same encoder architecture for our method and the baselines \citep{garnelo2018conditional, garnelo2018neural} in all experiments. For 1D and 2D functions, the data is fed in the form of input-output pairs $(x,y)$, where $x$ and $y$ are 1D values. We use MLPs with two hidden layer to encode the representations of these inputs. The number of hidden units in each layer is $d=50$. All MLPs have relu non-linearities except the final layer, which has no non-linearity.\\
\emph{Encoder}: Input(2) $\to $2 $\times$ (FC(50), ReLU) $\to$ FC(50).\\ While learning the representations of sinusoid functions with FCRL, we also scale the output scores with temperature to be $0.07$. We used the following hyper-parameter settings to train an encoder with FCRL.
\begin{figure*}%
	\centering
	\includegraphics[width=\textwidth]{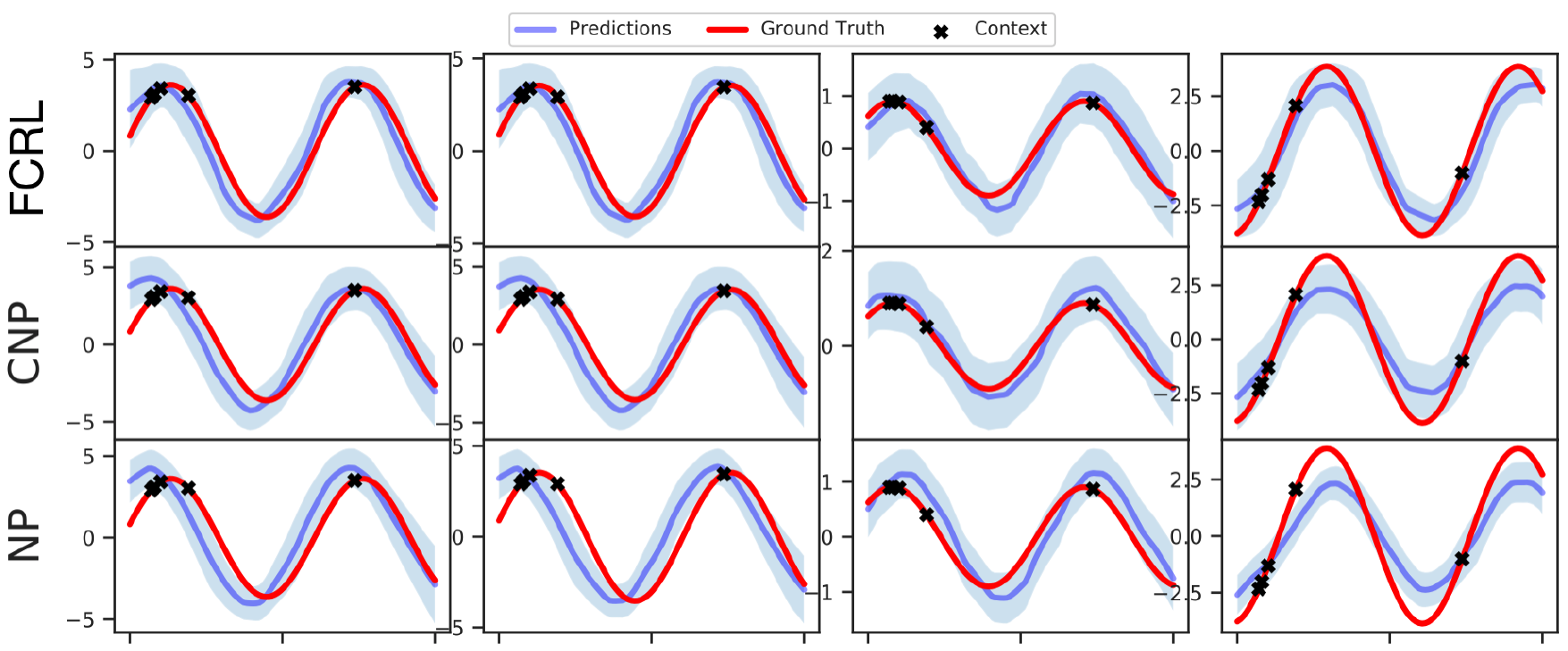}
	\caption{Additional results on 5-shot sinusoid regression. Each column corresponds to a different sinusoid function where only $5$ context points are given. The predictions of the decoder trained on FCRL based encoder are closer to the groundtruth.}
	\label{fig:sine_5_appendix}
\end{figure*}%
\paragraph{Downstream Tasks.}
To learn the subsequent task-specific decoders on the representations, we adapted the same data processing pipeline as above. For 1D functions, we train decoders for two different tasks: \emph{few-shot regression} and \emph{few-shot parameter identification}. The decoders for each task are trained with the same training dataset as was used to train the encoders. The training procedure for both downstream tasks on sinusoid functions is as follows

\begin{itemize}
    \item For Few-Shot Regression (FSR), we use an MLP architecture with two hidden layers. The same architecture are used in CNP \citep{garnelo2018conditional}, however in CNP the decoder and encoder are trained jointly. All the baselines and our model are trained for the same number of iterations.
    We used slightly higher learning rate to train the decoder as the training converges quite easily.\\
    \emph{FSR Decoder}: Input(50) $\to$ 2 $\times$ (FC(50), ReLU) $\to$ FC(1) .\\
    
    \item For Few-Shot Parameter Identification (FSPI), we train a linear decoder without any activation layers on the representations learned via FCRL and the baseline methods. The decoder is trained for only one epoch.\\
    \emph{FSPI Decoder}: Input(50) $\to$ FC(1) .\\
\end{itemize}

\begin{figure*}%
	\centering
	\includegraphics[width=\textwidth]{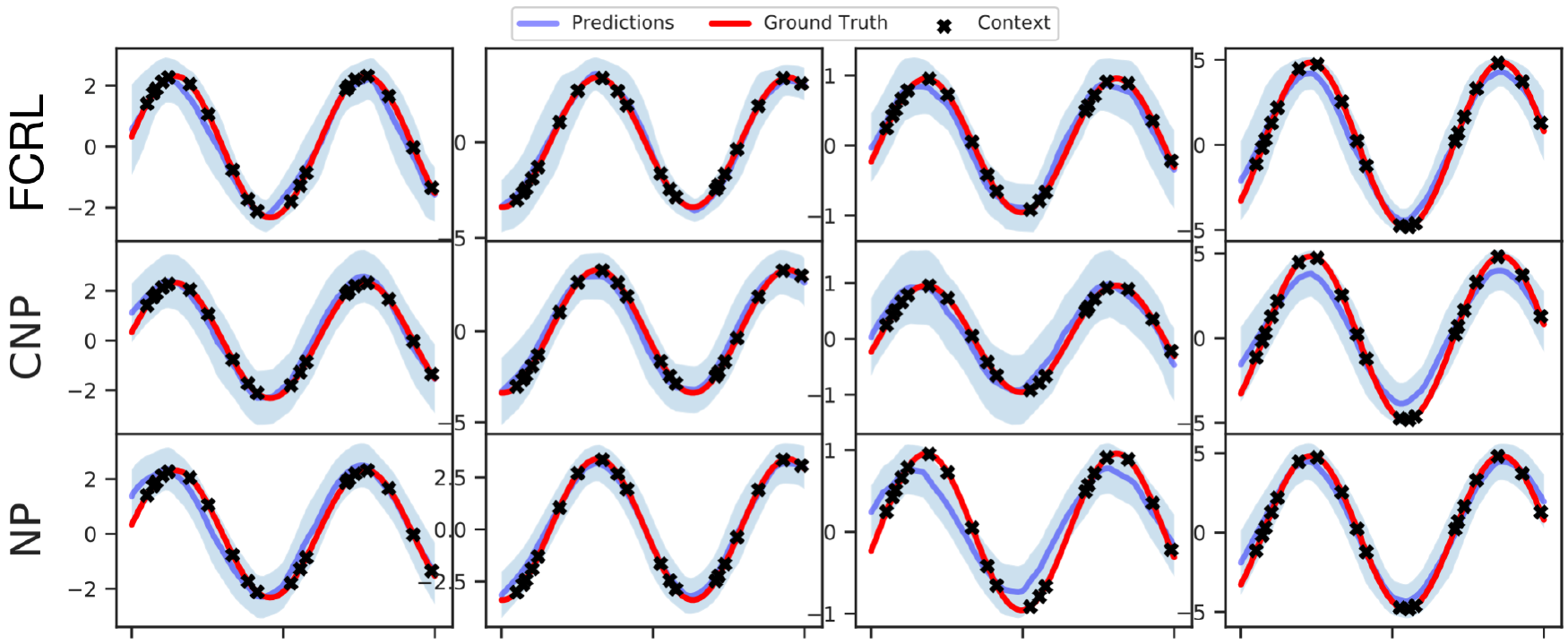}
	\caption{Additional results on 20-shot sinusoid regression. Each column corresponds to a different sinusoid function where only $20$ context points are given. The predictions of the decoder trained on FCRL based encoder are comparable to CNP and better than NP.}
	\label{fig:sine_20_appendix}
\end{figure*}

\paragraph{Additional Results.}
In \Cref{fig:sine_5_appendix} and \Cref{fig:sine_20_appendix}, we provide additional results on $5$-shot regression on test sets and compare the results with CNP and NP. The curves generated by the decoder using FCRL learned representations are closer to the groundtruth. The difference is evident in 5-shot experiments which supports the quantitative results in \Cref{tab:sinusoid}.

\section{Details of Experiments on 2D Functions}
\label{sec:exp-2D}
\paragraph{Implementation Details.}
In this experiment, we treat MNIST images as 2D functions. We adapt the architectures of encoders and decoders from the previous 1D experiments. However, due to the increased complexity of the function distributions we increase the number of hidden units of MLP to $d=128$. Moreover, the input $x$ is 2D as it corresponds to the cartesian coordinates of an image. The hyperparameter settings to train FCRL based encoder on such 2D function is given in \Cref{table:mnist_params}.
\begin{figure*}[h]
	\centering
	\includegraphics[width=\textwidth]{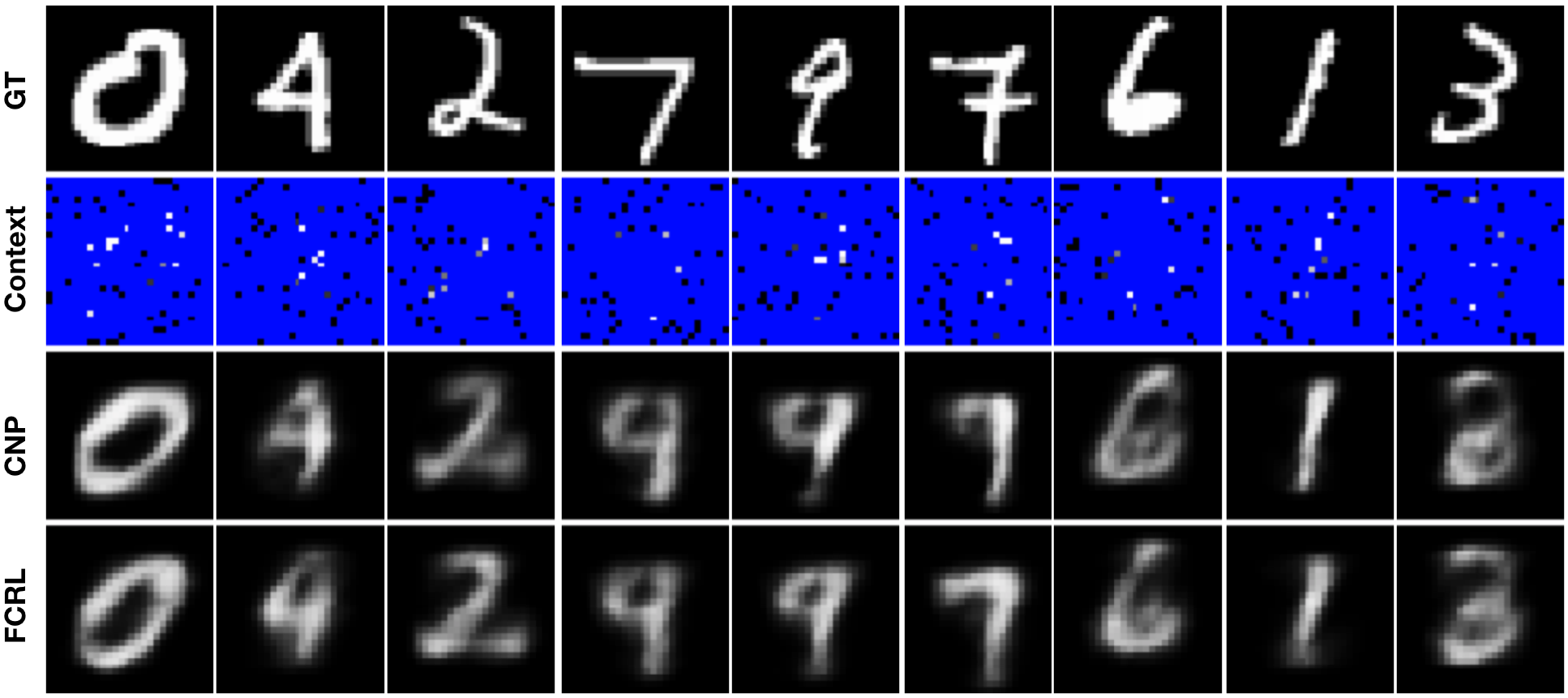}
	\caption{Additional results on 50-shot mnist image completion. The context is shown in the second row where target pixels are colored blue. Predictions made by a decoder trained on FCRL based encoder are slightly better than the CNP in terms of guessing the correct form of digits.}
	\label{fig:mnist_50_appendix}
\end{figure*}%

\begin{table*}[h!]\caption{Hyperparameters Settings for MNIST as 2D Functions Experiment.}\label{table:mnist_params}
\centering
\begin{subtable}[t]{0.5\linewidth}
\vspace{2mm}
    \begin{tabular}{l  l}
      \toprule
      \textbf{Parameter} & \textbf{Values}\\
      \midrule 
      Batch size & $16$\\
      Latent space dimension & $128$\\
      Temperature: $\tau$ & 0.007 \\
      Number of subsets: $J$ & 40 \\
      Max number of context points: $N$ & 200\\
      Epochs & 100 \\
      Critic & Nonlinear \\
      Optimizer & Adam\\
      Adam: beta1 & 0.9\\
      Adam: beta2 & 0.999\\
      Adam: epsilon & 1e-8\\
      Adam: learning rate & 0.0006\\
      Learning Rate Scheduler & Cosine \\ 
      \bottomrule
    \end{tabular}
    \vspace{0.5cm}
    \caption{Hyperparameters to train FCRL based encoder for 2D functions.}
\end{subtable}%
\hspace{5mm}
\begin{subtable}[t]{0.4\linewidth}
\vspace{2mm}
    \begin{tabular}{l  l}
      \toprule
      \textbf{Parameter} & \textbf{Values}\\
      \midrule 
      Batch size & $16$\\
      Epochs & 100 \\
      Critic & Nonlinear \\
      Optimizer & Adam\\
      Adam: beta1 & 0.9\\
      Adam: beta2 & 0.999\\
      Adam: epsilon & 1e-8\\
      Adam: learning rate & 0.001\\
      Learning Rate Scheduler & Cosine \\ 
      \bottomrule
    \end{tabular}
    \vspace{0.5cm}
    \caption{Hyperparameters to train Few-Shot Image Completion (FSIC) Decoder trained on FCRL learned representations.}
\end{subtable}%
\end{table*}
\subsection{Downstream Tasks.} We consider two downstream tasks to evaluate the quality of the representations learned on 2D functions: \emph{few-shot image completion} and \emph{few-shot content classification}. A separate decoder is trained for both of these tasks.
\begin{itemize}
    \item For Few-Shot Image Completion (FSIC), we use an MLP based decoder with two hidden layers. The decoder is trained on the same training data for the same number of iterations. Details are given in \Cref{table:mnist_params}(b).
    \item For Few-Shot Content Classification (FSCC), we train a linear regression on top of the representations obtained by both FCRL and the baselines. The decoder is trained for only one epoch.
\end{itemize}

\begin{figure*}[h]
	\centering
	\includegraphics[width=\textwidth]{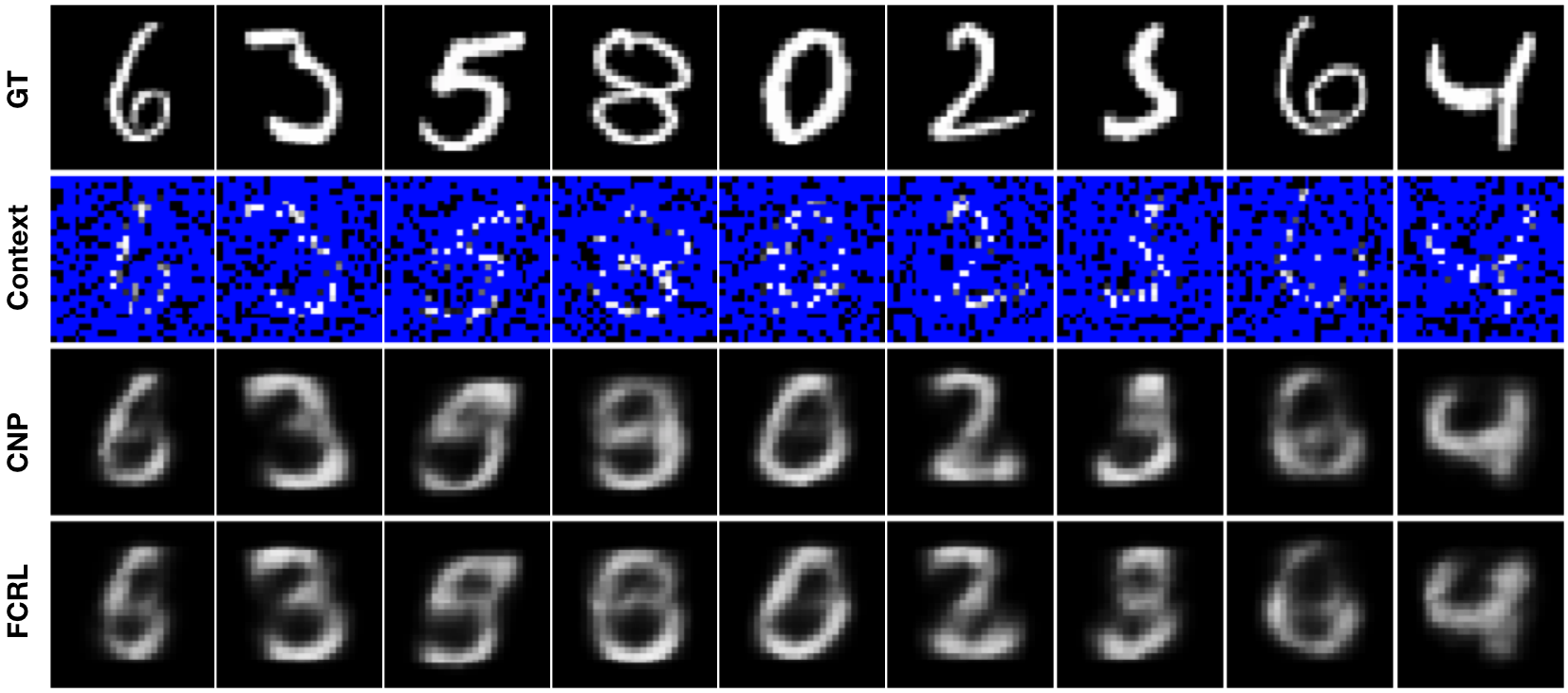}
	\caption{Additional results on 200-shot mnist image completion. The context is shown in the second row where target pixels are colored blue. Predictions made by a decoder trained on FCRL based encoder are comparable to CNP.}
	\label{fig:mnist_200_appendix}
\end{figure*}

\paragraph{Additional Results.}
In \Cref{fig:mnist_50_appendix} and \Cref{fig:mnist_200_appendix}, we provide additional results for $50$-shot and $200$-shot image completion. We can see that the the results from FCRL based decoder consistently perform better than CNP in low-shot scenario of $50$ context points. In $200$-shot scenario, the results look comparable to CNP.

\end{document}